\documentclass[11pt,a4paper]{article}
\usepackage{times,latexsym}
\usepackage{url}
\usepackage[T1]{fontenc}

\usepackage[table,xcdraw,dvipsnames]{xcolor}
\usepackage{amsmath}
\usepackage{amssymb}
\usepackage{mathtools}
\usepackage{amsthm}
\usepackage{bbm}
\usepackage{bm}
\usepackage{mathtools}

\newcommand{\minus}{\kern-0.07em-\kern-0.07em}
\newcommand{\indicator}{\kern-0.16em\raisebox{-0.22em}{\scalebox{1.5}{$\mathbbm{1}$}}\kern-0.16em}

\usepackage{microtype}
\usepackage{booktabs}
\usepackage{multirow}
\usepackage{pifont}
\usepackage{xspace}
\usepackage{enumitem}
\usepackage{arydshln}
\usepackage{algorithm}
\usepackage{algorithmic}
\usepackage{hhline}
\usepackage{wrapfig}

\definecolor{skyblue}{RGB}{173, 216, 230}  
\definecolor{lightred}{RGB}{255, 182, 193} 
\definecolor{darkgreen}{RGB}{34, 139, 34}  
\definecolor{darkpink}{RGB}{199, 21, 133}  

\theoremstyle{plain}
\newtheorem{theorem}{Theorem}[section]

\theoremstyle{definition}
\newtheorem{definition}[theorem]{Definition}

\theoremstyle{remark}

\usepackage{times}
\usepackage{latexsym}

\usepackage[T1]{fontenc}

\usepackage[utf8]{inputenc}

\usepackage{microtype}

\usepackage{inconsolata}

\usepackage{graphicx}
\usepackage[textsize=tiny]{todonotes}

\usepackage{amsfonts} 
\usepackage{empheq} 
\usepackage{mathrsfs}
\usepackage{ragged2e}
\usepackage{listings}

\usepackage[acceptedWithA]{tacl2021v1}
\usepackage{tacl2021v1}
\usepackage[capitalize,noabbrev]{cleveref}

\newcommand{\cmark}{\textcolor{green}{\ding{51}}}   
  
\newcommand{\alg}{\textbf{\texttt{\textbf{DReaMAD}}}\xspace}

\usepackage{xparse}
\usepackage{tabularx}
\usepackage{booktabs}
\usepackage{enumitem}
\usepackage{array}

\usepackage{xspace,mfirstuc,tabulary}

\newif\iftaclinstructions
\taclinstructionsfalse 
\iftaclinstructions
\renewcommand{\confidential}{}
\renewcommand{\anonsubtext}{(No author info supplied here, for consistency with
TACL-submission anonymization requirements)}
\newcommand{\instr}
\fi

\iftaclpubformat 

\else

\fi

\title{From Belief Entrenchment to Robust Reasoning in LLM Agents}

\author{Jihwan Oh{\textsuperscript{1$*$}} \quad Minchan Jeong{\textsuperscript{1$*$}} \quad Jongwoo Ko{\textsuperscript{2$\dagger$}} \quad Se-Young Yun {\textsuperscript{1$\dagger$}} \\
{\textsuperscript{1}}KAIST AI\quad{\textsuperscript{2}}Microsoft\\ 
\texttt{\{ericoh929, mcjeong\}@kaist.ac.kr}
}

\date{}

\begin{document}

\maketitle
\begingroup
\renewcommand{\thefootnote}{} %
\footnotetext{%
  \textsuperscript{*}Equal contribution.\hspace{1em}%
  \textsuperscript{$\dagger$}Corresponding authors.%
}
\endgroup
\begin{abstract}
Multi-Agent Debate (MAD) has emerged as a promising inference scaling method for Large Language Model (LLM) reasoning. However, it frequently suffers from belief entrenchment, where agents reinforce shared errors rather than correcting them. Going beyond merely identifying this failure, we decompose it into two distinct root causes: (1) the model's biased \textit{static initial belief} and (2) \textit{homogenized debate dynamics} that amplify the majority view regardless of correctness. To address these sequentially, we propose \alg (\textbf{D}iverse \textbf{Rea}soning via \textbf{M}ulti-\textbf{A}gent \textbf{D}ebate). Our framework first rectifies the static belief via strategic prior knowledge elicitation, then reshapes the debate dynamics by enforcing perspective diversity. Validated on our new \textit{MetaNIM Arena} benchmark, \alg significantly mitigates entrenchment, achieving a +9.5\% accuracy gain over ReAct prompting and a +19.0\% higher win rate than standard MAD.
\end{abstract}

\section{Introduction}
\label{sec:intro}

While Large Language Models (LLMs) have shown impressive problem-solving capabilities~\citep{achiam2023gpt, dubey2024llama}, their ability to self-correct remains inconsistent. Single-agent mechanisms like Self-Refinement~\citep{madaan2024self} or Self-Consistency~\citep{wang2022self} often degrade performance when models cannot accurately assess their own reasoning~\citep{huang2024large}. To address this, Multi-Agent Debate (MAD; \citealt{chan2023chateval, du2023improving, liang2023encouraging}) has emerged as a promising paradigm for inference-time scaling, leveraging inter-agent critique to refine outputs.

\begin{figure*}[ht]
\vspace{5pt}
\begin{center}
\centerline{\includegraphics[width=0.95\textwidth]{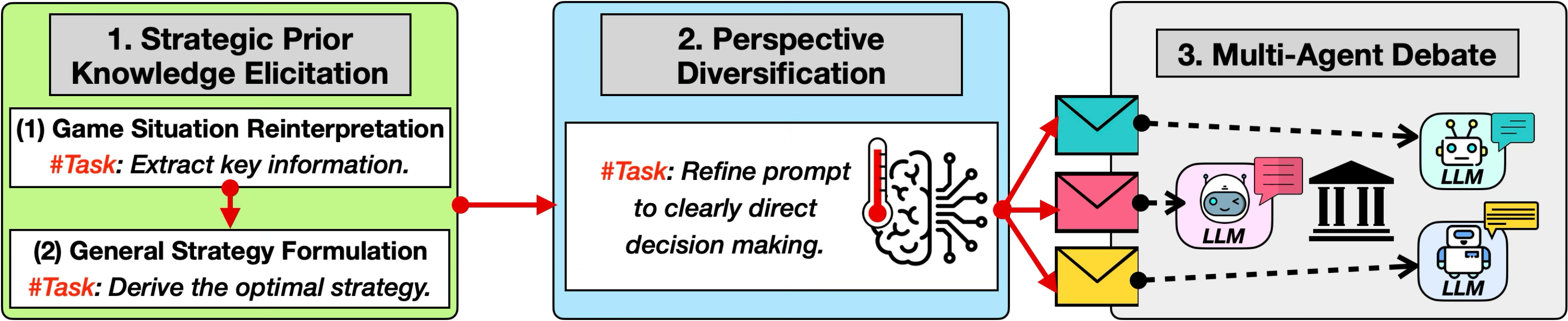}}
\vspace{-3pt}
\caption{\alg framework. \alg improves LLM reasoning by combining \textbf{Strategic Prior Knowledge Elicitation} and \textbf{Perspective Diversification}. First, the model reinterprets the problem and formulates high-level strategies to reduce belief entrenchment. In the second stage, multiple agents adopt distinct viewpoints, engage in structured debate, and refine their conclusions to enhance decision-making.}
\label{fig:methods}
\end{center}
\vspace{-25pt}
\end{figure*}

However, we observe that standard MAD is brittle in sequential, strategic environments, where a single suboptimal decision triggers consecutive failures that the consensus process fails to correct.
Rather than challenging errors, homogeneous agents frequently converge into an ``echo chamber,'' amplifying inherent cognitive biases.
As we demonstrate in Section~\ref{sec:analysis}, this failure stems from a debate dynamic that prioritizes initial response frequency over logical validity, effectively rewarding popularity rather than correctness. Unlike static tasks~\citep{cobbe2021training, edwards1994portable, he2020box} where errors can be masked, strategic games enforce a rigorous penalty for every suboptimal reasoning step, making them ideal for dissecting cascading failures.

To address these limitations, we introduce \textit{MetaNIM Arena}, a rigorous benchmark for adversarial strategic decision-making. Using this environment, we identify a critical failure mode termed \textbf{belief entrenchment}, where interaction actually hinders error correction. We decompose this phenomenon into two root causes:
\begin{enumerate}[label=\textbf{(\arabic*)}, leftmargin=*, itemsep=1pt] 
\item \textbf{Static Initial Belief:} The model's inherent bias prior to interaction, where it relies on immediate context rather than strategic foresight.
\item \textbf{Homogenized Debate Dynamics:} A self-reinforcing mechanism where the persuasiveness of an argument is dictated by its initial popularity among homogeneous agents, suppressing valid but less probable reasoning paths.
\end{enumerate}

Building on the \textit{Learning from Multiple Approaches} framework~\citep{national2005students}, which suggests that engaging with multiple problem-solving representations mitigates bias, we propose \alg (\textbf{D}iverse \textbf{Rea}soning via \textbf{M}ulti-\textbf{A}gent \textbf{D}ebate with Refined Prompt). As shown in Figure~\ref{fig:methods}, \alg addresses the identified root causes by (1) eliciting strategic prior knowledge to correct static belief and (2) systematically modifying prompts to foster diverse perspectives, thereby reshaping debate dynamics. Our key contributions are as follows:
\vspace{-5pt}
\begin{itemize}[leftmargin=*, itemsep=0pt]
    \item We identify and decompose \textbf{belief entrenchment} in Multi-Agent Debate, showing that standard MAD acts as a dynamic amplifier of static cognitive biases.
    \item We propose \alg, a novel framework that sequentially targets the root causes of belief entrenchment through strategic knowledge elicitation and perspective diversification. It achieves a +9.5\% accuracy gain over ReAct prompting on the \textit{MetaNIM Arena} dataset and a +19.0\% higher win rate than MAD in the simulator.
    \item We introduce \textbf{MetaNIM Arena}, a benchmark designed to evaluate LLMs in adversarial strategic decision-making, enabling precise assessment of reasoning robustness and strategic adaptability.
    \vspace{-5pt}
\end{itemize}
\section{MetaNIM Arena}
\label{sec:metanim}

We introduce \textit{MetaNIM Arena}, a benchmark designed to rigorously evaluate adversarial strategic reasoning in LLMs. Unlike static QA tasks, this benchmark requires agents to make sequential decisions in dynamic environments where optimal play is mathematically definable but non-trivial to execute.

\subsection{Mathematical Foundation: Impartial Games}
The arena consists of five \textit{impartial games}: NIM, Fibonacci Nim, Kayles, Chomp, and Corner Queen. These games share key properties: (1) \textbf{Full Observability:} The entire game state is visible to both players; (2) \textbf{Impartiality:} Available moves depend solely on the state, not on which player is moving; (3) \textbf{Termination:} The game is guaranteed to end in a finite number of moves. 

\begin{figure*}[t!] 
\begin{center} 
\centerline{\includegraphics[width=0.95\textwidth]{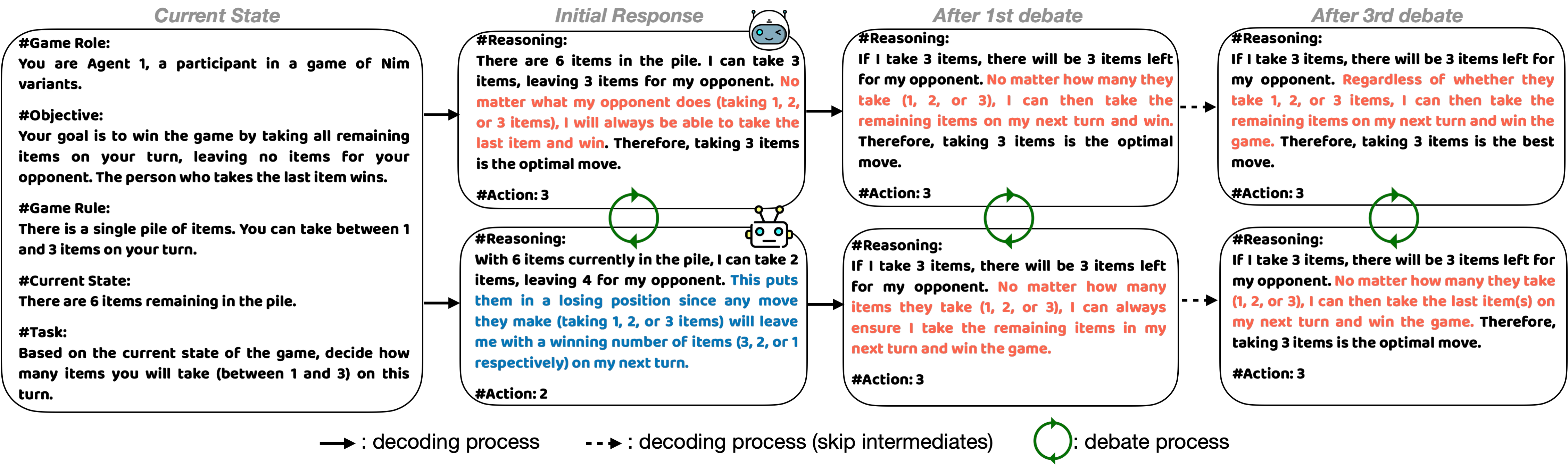}}
\vspace{-5pt}
\caption{An example demonstrating how the debate process converges to a biased outcome. We observed that belief entrenchment occurs in the first debate. \textcolor{blue}{Blue text indicates the correct reasoning} and \textcolor{orange}{orange text indicates the strong consistent (biased) reasoning}. The second debate is omitted, as its procedure replicates the first and third; all debates use \texttt{GPT-4o-mini} as the debating agent.}
\label{fig:debate_process}
\end{center}
\vspace{-25pt}
\end{figure*}

Crucially, for any non-terminal state in these games, one player holds a provably winning strategy assuming optimal play. This is grounded in the \textbf{Sprague-Grundy Theorem}, which maps every impartial game state to a non-negative integer called a \textit{Grundy number} (or \textit{nimber}). A state with a Grundy number of 0 is a losing position (P-position), while any non-zero state is a winning position (N-position). This provides a clear, objective criterion for optimality: an agent reasoning correctly should always move the game to a state with a Grundy number of 0. 

\subsection{Game Variants and Complexity}
\textit{MetaNIM Arena} includes diverse game settings to test different aspects of strategic reasoning:
\begin{itemize}[leftmargin=*, itemsep=0pt]
    \item \textbf{NIM:} The foundational game where players remove objects from heaps. The winning strategy relies on the XOR sum of heap sizes (Nim-sum).
    \item \textbf{Fibonacci Nim:} A dynamic variant where the maximum removal is constrained by the opponent's previous move (up to $2\times$). This introduces complex state dependencies requiring forward lookahead.
    \item \textbf{Kayles:} Players knock down pins, potentially splitting a row into two independent subgames. This tests the agent's ability to decompose problems and analyze disjoint game components.
    \item \textbf{Chomp:} A geometric game on a grid where players consume blocks. While a winning strategy is guaranteed to exist, no simple closed-form solution is known for general grid sizes, making it a test of heuristic search and pattern recognition.
    \item \textbf{Corner Queen:} A spatial game equivalent to \textit{Wythoff's game}, where winning positions follow the Beatty sequence (related to the Golden Ratio), requiring precise numerical reasoning.
\end{itemize}
We provide a more detailed theoretical explanation for impartial games and each game variant in Appendix~\ref{app:sec:comb_games}.

\subsection{Evaluation Modes}
We utilize \textit{MetaNIM Arena} in two complementary modes:

\paragraph{Interactive Simulator.}
This mode evaluates an agent's adaptability in a turn-based environment. The agent plays against a strong adaptive opponent powered by \texttt{GPT-4o} using the ReAct framework~\citep{yao2023react}. To ensure rigorous evaluation, the agent is always initialized in a winning position (N-position). Performance is measured by win rate, reflecting the agent's ability to maintain the winning advantage against an optimal or near-optimal adversary. We employ various configurations (e.g., different board sizes, starting items) as detailed in Appendix~\ref{app:sec:dataset}.

\paragraph{Static Dataset.}
To analyze reasoning quality at a granular level, we constructed a static dataset of critical game states. Each entry consists of a state description and its unique optimal move (calculated via Grundy numbers). This allows us to measure \textbf{Decision Accuracy}, defined as the percentage of turns where the agent selects the theoretically optimal move, without the noise of full game simulations.
\section{Analyzing Belief Entrenchment in Debate}
\label{sec:analysis}

Before introducing our method, we systematically analyze why standard Multi-Agent Debate (MAD) often fails to correct errors and instead reinforces inherent belief. We utilize the \textit{MetaNIM Arena} environment to quantitatively measure this phenomenon.

\subsection{Experimental Setup and Definitions}
\label{subsec:setup}

To rigorously analyze the debate dynamics, we focus on the \textit{Fibonacci Nim} game, where the optimal move is mathematically deterministic. This allows us to clearly distinguish between the model's inherent preference and the ground-truth optimal action.  

\paragraph{Estimating Argument Probability.}
Departing from token-level log-probabilities used in prior work, we estimate the probability $p_i$ of an argument (action) $i$ via empirical generation frequency. Specifically, we sample $N=20$ responses at a fixed temperature ($T=0.7$) to determine the action distribution. Note that $p_i$ represents the model's belief in a specific context; it is not static but shifts based on the prompting strategy.

\paragraph{Defining Belief and Consistency.}
We conceptualize the model's \textit{static belief} as its initial action probability distribution, denoted as $p_i^{(0)}$, prior to any multi-agent interaction.
\begin{itemize}
    \item \textbf{Optimal Action:} The theoretically unique move that guarantees a winning strategy from the current state, serving as the ground truth for correctness.
    \item \textbf{Suboptimal Tendency:} A systematic preference for a losing action driven by heuristic shortcuts.
    \item \textbf{Strong Consistency:} A state where the model's belief is highly concentrated, specifically when the dominant action's probability exceeds the majority threshold ($p_i > 0.5$). This indicates a solidified stance likely to dominate the consensus process.
\end{itemize}

\subsection{Empirical Evidence and Modeling}
\label{subsec:empirical_evidence}

\begin{figure}[t]
    \centering
    \includegraphics[width=1.0\columnwidth]{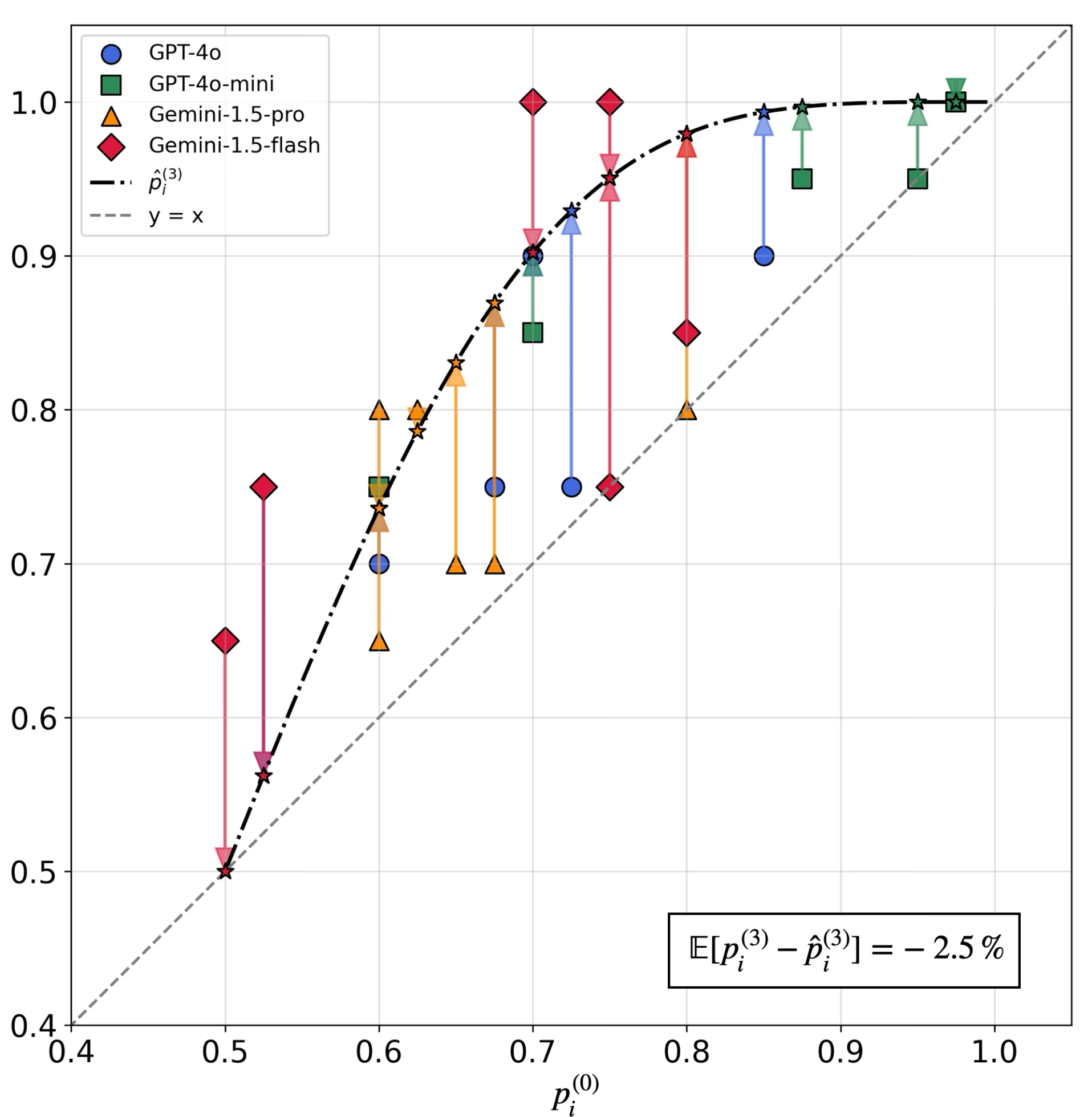} %
    \vspace{-15pt}
    \caption{Belief entrenchment across models: showing that even after the debate concludes, strongly consistent actions continue to exhibit strong consistency, reinforcing biased action distributions in the Fibonacci game.}
    \label{fig:belief_entrenchment}
    \vspace{-15pt}
\end{figure}

\vspace{-5pt}
\begin{figure}[t]
    \centering    
    \includegraphics[width=1.0\columnwidth]{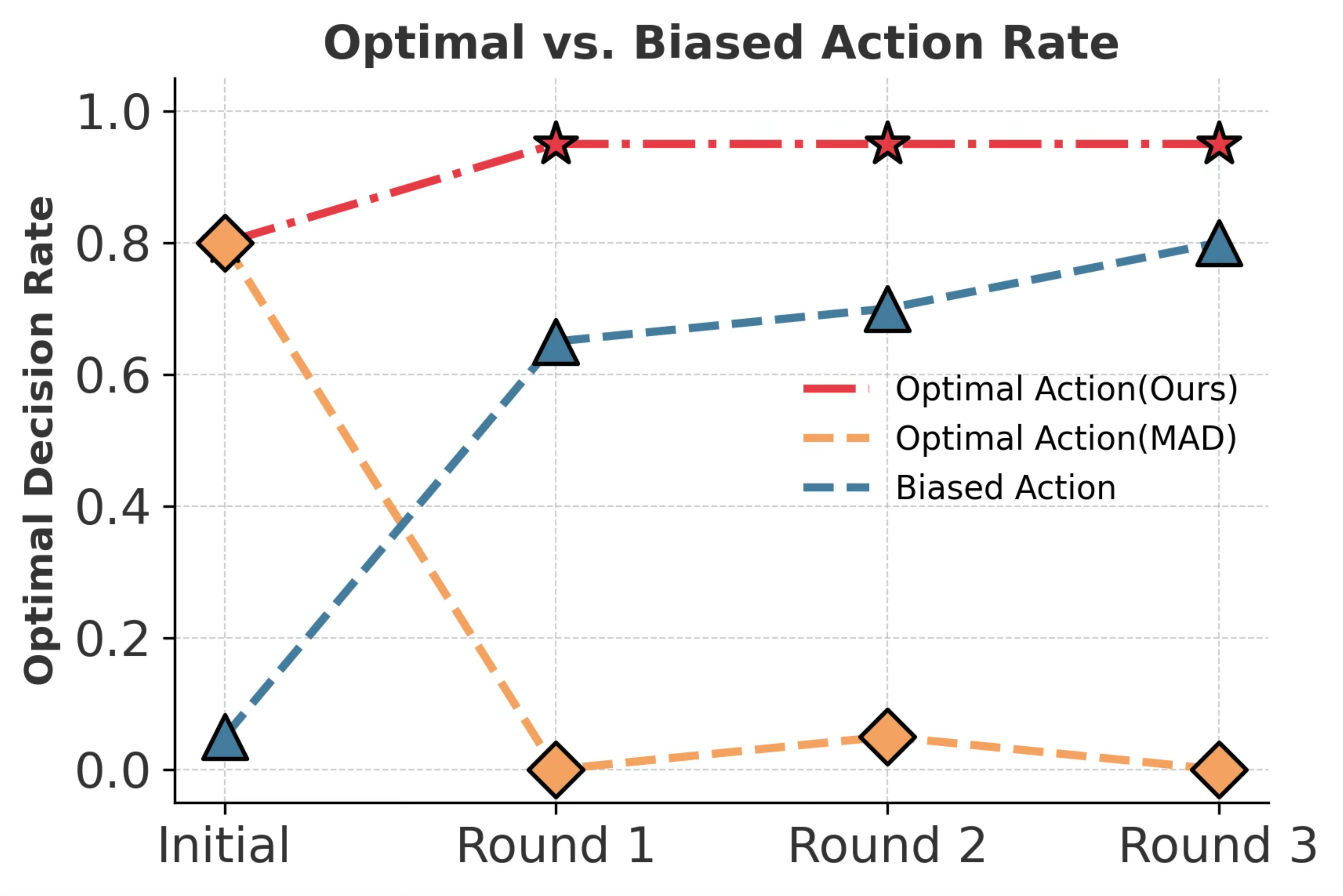} 
        \vspace{-5pt}
    \caption{Belief Entrenchment in Action (NIM). Despite one agent proposing the optimal move, the debate converges to the suboptimal majority view, illustrating how valid reasoning is suppressed by the model's static belief.}
    \label{fig:motif_example}
    \vspace{-5pt}
\end{figure}

We conducted experiments using 3 rounds of standard MAD \cite{du2023improving} with various agents against \texttt{GPT-4o}. Figure~\ref{fig:belief_entrenchment} illustrates the change in probability of the dominant action after the debate.

\paragraph{Observation: Belief Entrenchment.}
As shown in Figure~\ref{fig:belief_entrenchment}, actions that initially exhibited Strong Consistency ($p^{(0)} > 0.5$) consistently increased their probability after debate ($p^{(3)} > p^{(0)}$), regardless of whether the action was optimal or suboptimal. Furthermore, all data points for Strong Consistency lie above the $y=x$ line. This empirically demonstrates \textbf{belief entrenchment}: the debate process acts as an amplifier for the majority opinion, rather than a correction mechanism based on logical validity, as illustrated in Figure~\ref{fig:debate_process}. This leads agents to converge on a suboptimal consensus, as shown in Figure~\ref{fig:motif_example}, where a correct counterargument is overridden by the reinforced bias, leading the agents to converge on a suboptimal consensus.

\begin{figure*}[t]
\begin{center}
\centerline{\includegraphics[width=1.0\textwidth]{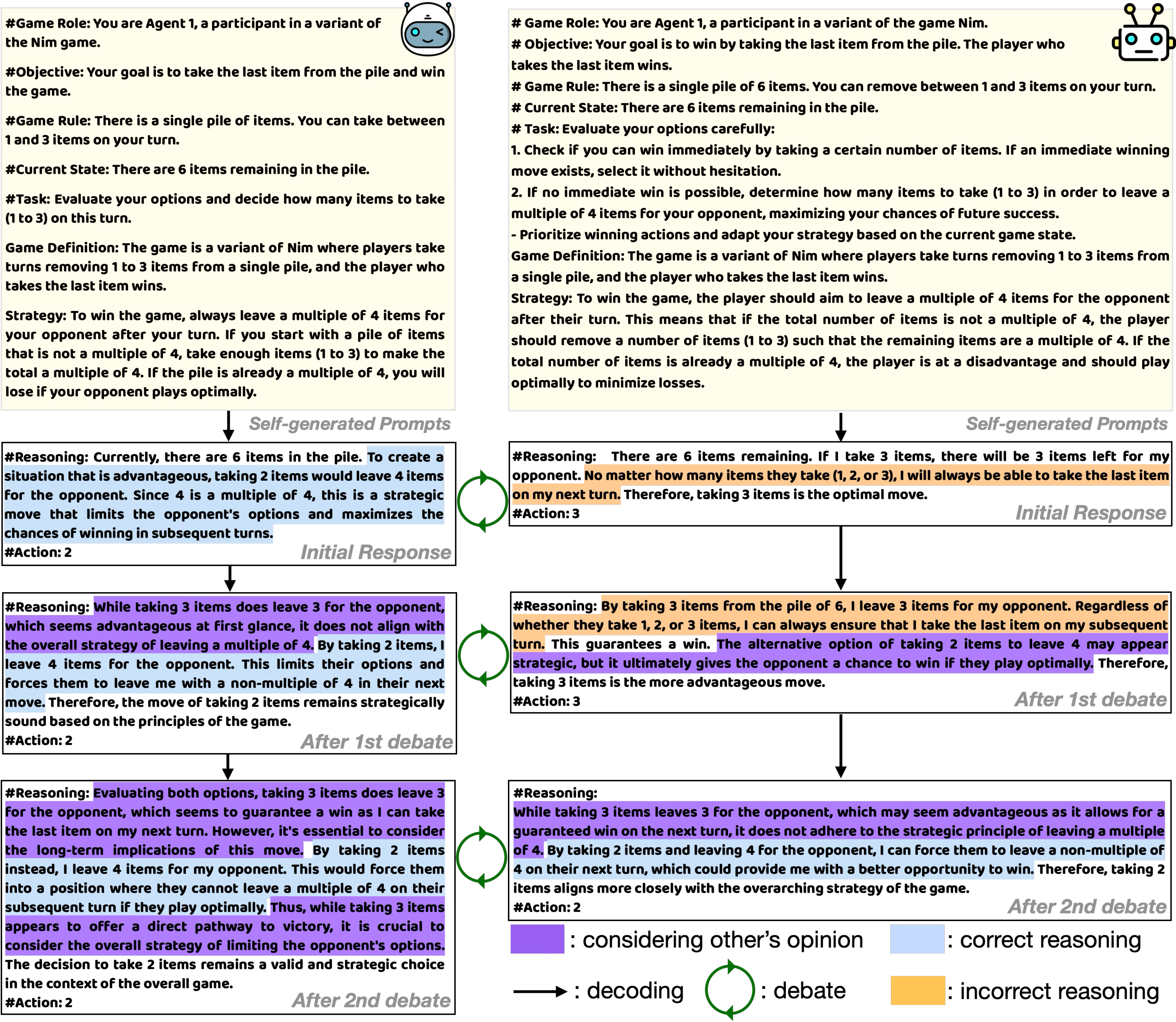}}
\caption{In this example, we illustrate how the debate process converges to an optimal outcome using our algorithm, \alg. We begin with the same \textit{current state} shown in Figure~\ref{fig:debate_process}, employing self-generated prompts for each LLM agent.}
\label{fig:dreamad_process}
\end{center}
\vspace{-20pt}
\end{figure*}

\paragraph{Modeling the Dynamics.}
To explain the mechanism behind this phenomenon, we propose a probabilistic interaction model based on \textbf{pairwise agent debate}. We posit that the persuasiveness of an argument is not determined by its logical validity, but is proportional to the model's static initial belief, $p_i^{(0)}$. This implies that agents are more likely to accept arguments that align with their inherent biases.

Mathematically, the probability of action $i$ at round $t+1$, denoted as $p^{(t+1)}_i$, evolves as follows:
\begin{equation}
\label{eq:prob_update}
p^{(t+1)}_i = \underbrace{{p^{(t)}_i} \cdot {p^{(t)}_i} }_{\mathclap{\text{Consensus}}} \mkern2mu + \mkern9mu \underbrace{2 p^{(t)}_i (1-p^{(t)}_i) \cdot p^{(0)}_i}_{\mathclap{\text{Belief-Driven Resolution}}}\,,
\end{equation}
where the first term represents the probability of both agents agreeing on action $i$, and the second term models conflict resolution biased by the initial popularity $p_i^{(0)}$.

Our simple model generates the theoretical curve in Figure~\ref{fig:belief_entrenchment}, which closely fits the empirical data points with an average residual of just $-2.5\%$. This tight alignment validates our hypothesis: the debate dynamics act as an amplifier of static bias ($p^{(0)}$) rather than a filter for reasoning quality.

\paragraph{Decomposing the Failure.}
Crucially, the strong fit of our model allows us to \textbf{decompose} belief entrenchment into two distinct root causes, motivating our proposed method:
\begin{enumerate}
    \item \textbf{Static Initial Belief ($p^{(0)}$):} The model's inherent bias prior to any interaction. If the initial probability of the correct action is low ($p^{(0)}_{\mathrm{correct}} < 0.5$), the debate cannot recover it.
    \item \textbf{Homogenized Debate Dynamics:} The functional dependency where an argument's persuasiveness is dictated by its initial likelihood ($p^{(0)}$) rather than its logical validity. As evidenced by our model fit, this creates an echo chamber where homogeneous agents reinforce the dominant view regardless of its correctness.
\end{enumerate}
\section{{\alg}: Counteracting Belief Entrenchment}
\label{sec:method}

\begin{table*}[t!]
    \centering
    \scalebox{0.90}{
    \footnotesize 
    \begin{tabular}{ccccccc}
        \toprule
        \textbf{LLM Models} & \textbf{Prompting Methods} & \textbf{NIM} & \textbf{Fibonacci} & \textbf{Chomp} & \textbf{Kayles} & \textbf{Average} \\
        \midrule
        \multirow{3}{*}{\texttt{GPT-4o}} & ReAct & 0.95 \scriptsize $\pm$ 0.04 & 0.33 \scriptsize $\pm$ 0.04 & 0.18 \scriptsize $\pm$ 0.07 & 0.19 \scriptsize $\pm$ 0.08 & 0.41 \\ 
        &+ CoT-Prompting & 0.96 \scriptsize $\pm$ 0.04 & 0.43 \scriptsize $\pm$ 0.11 & \textbf{0.28} \scriptsize $\pm$ 0.09 & 0.20 \scriptsize $\pm$ 0.10 & \textbf{0.47} \\
        &\cellcolor{gray!20} \alg$^{\kern-0.3em(-)}$ & \cellcolor{gray!20} \textbf{0.98} \scriptsize $\pm$ 0.04 & \cellcolor{gray!20} \textbf{0.44} \scriptsize $\pm$ 0.09 & \cellcolor{gray!20} 0.23 \scriptsize $\pm$ 0.10 & \cellcolor{gray!20} \textbf{0.23} \scriptsize $\pm$ 0.12 & \cellcolor{gray!20} \textbf{0.47} \\[-0.18em]
        \midrule
         \multirow{3}{*}{\texttt{GPT-4o-mini}} & ReAct & 0.75 \scriptsize $\pm$ 0.05 & 0.33 \scriptsize $\pm$ 0.04 & 0.40 \scriptsize $\pm$ 0.07 & 0.12 \scriptsize $\pm$ 0.06 & 0.40 \\
        & + CoT-Prompting & 0.84 \scriptsize $\pm$ 0.08 & 0.36 \scriptsize $\pm$ 0.06 & 0.61 \scriptsize $\pm$ 0.05 & 0.02 \scriptsize $\pm$ 0.03 & 0.46 \\
        & \cellcolor{gray!20} \alg$^{\kern-0.3em(-)}$ & \cellcolor{gray!20} \textbf{1.00} \scriptsize $\pm$ 0.00 & \cellcolor{gray!20} \textbf{0.49} \scriptsize $\pm$ 0.17 & \cellcolor{gray!20} \textbf{0.62} \scriptsize $\pm$ 0.10 & \cellcolor{gray!20}  \textbf{0.18} \scriptsize $\pm$ 0.11 & \cellcolor{gray!20} \textbf{0.57}  \\[-0.18em]
        \midrule
         \multirow{3}{*}{\texttt{Gemini-1.5-pro}} & ReAct & 0.82 \scriptsize $\pm$ 0.06 & 0.42 \scriptsize $\pm$ 0.04 & 0.19 \scriptsize $\pm$ 0.08 & 0.57 \scriptsize $\pm$ 0.04 & 0.50 \\
         &+ CoT-Prompting & 0.88 \scriptsize $\pm$ 0.05 & 0.47 \scriptsize $\pm$ 0.11 & 0.22 \scriptsize $\pm$ 0.11 & 0.59 \scriptsize $\pm$ 0.04 & 0.54 \\
        & \cellcolor{gray!20} \alg$^{\kern-0.3em(-)}$ & \cellcolor{gray!20} \textbf{0.97} \scriptsize $\pm$ 0.04 & \cellcolor{gray!20} \textbf{0.53} \scriptsize $\pm$ 0.07 & \cellcolor{gray!20} \textbf{0.24} \scriptsize $\pm$ 0.05 & \cellcolor{gray!20} \textbf{0.72} \scriptsize $\pm$ 0.12 & \cellcolor{gray!20} \textbf{0.62} \\[-0.18em] \midrule
        \multirow{3}{*}{\texttt{Gemini-1.5-flash}} & ReAct & 0.94 \scriptsize $\pm$ 0.02 & 0.35 \scriptsize $\pm$ 0.04 & 0.05 \scriptsize $\pm$ 0.03 & 0.01 \scriptsize $\pm$ 0.02 & 0.34 \\
        & + CoT-Prompting & 0.93 \scriptsize $\pm$ 0.02 & 0.33 \scriptsize $\pm$ 0.07 & \textbf{0.09} \scriptsize $\pm$ 0.04 & 0.0 \scriptsize $\pm$ 0.00 & 0.34 \\
        & \cellcolor{gray!20} \alg$^{\kern-0.3em(-)}$ & \cellcolor{gray!20} \textbf{0.97} \scriptsize $\pm$ 0.04 & \cellcolor{gray!20} \textbf{0.45} \scriptsize $\pm$ 0.06 & \cellcolor{gray!20} 0.05 \scriptsize $\pm$ 0.00 & \cellcolor{gray!20} \textbf{0.42} \scriptsize $\pm$ 0.06 & \cellcolor{gray!20} \textbf{0.46} \\[-0.18em]
        \midrule
        \multirow{3}{*}{\texttt{Qwen3-4B}} 
        & ReAct 
        & \textbf{0.99} \scriptsize $\pm$ 0.02 
        & \textbf{0.42} \scriptsize $\pm$ 0.09 
        & 0.08 \scriptsize $\pm$ 0.05 
        & 0.47 \scriptsize $\pm$ 0.03 
        & \textbf{0.49} \\
        & + CoT-Prompting 
        & 0.98 \scriptsize $\pm$ 0.02 
        & 0.38 \scriptsize $\pm$ 0.04 
        & \textbf{0.08} \scriptsize $\pm$ 0.03 
        & \textbf{0.48} \scriptsize $\pm$ 0.02 
        & 0.48 \\
        & \cellcolor{gray!20} \alg$^{\kern-0.3em(-)}$ 
        & \cellcolor{gray!20} \textbf{0.99} \scriptsize $\pm$ 0.02 
        & \cellcolor{gray!20} \textbf{0.42} \scriptsize $\pm$ 0.09 
        & \cellcolor{gray!20} 0.04 \scriptsize $\pm$ 0.02 
        & \cellcolor{gray!20} 0.47 \scriptsize $\pm$ 0.06 
        & \cellcolor{gray!20} 0.48 \\[-0.18em]
        \midrule
        \multirow{3}{*}{\textbf{Average}} & ReAct & 0.89 & 0.37 & 0.18 & 0.27 & - \\
        & + CoT-Prompting & 0.92 & 0.39 & \textbf{0.26} & 0.26 & - \\
        & \cellcolor{gray!20} \alg$^{\kern-0.3em(-)}$ & \cellcolor{gray!20} \textbf{0.98} & \cellcolor{gray!20} \textbf{0.47} & \cellcolor{gray!20} 0.24 & \cellcolor{gray!20} \textbf{0.40} & \cellcolor{gray!20} - \\[-0.18em]
        \bottomrule
    \end{tabular}
    }%
    \caption{Effect of Strategic Prior Knowledge Elicitation module. \alg$^{\kern-0.3em(-)}$ indicates our method except multi-agent debate process. We can fully evaluate reasoning ability between different prompting methods. The metric accuracy of selecting optimal action is used. The best results are highlighted in \textbf{bold}.}
    \label{tab:reasoning_ability}
    \vspace{-5pt}
\end{table*}

To address the limitations of standard MAD identified in Section~\ref{sec:analysis}, we introduce \alg (Diverse Reasoning via Multi-Agent Debate with Refined Prompt). Our framework counteracts belief entrenchment by targeting its two root causes: the initial static bias ($p^{(0)}$) and the homogenized debate dynamics.

\subsection{Stage 1: Strategic Prior Knowledge Elicitation (SPKE)}
The first stage aims to correct the model's initial reasoning bias before any debate begins. Standard prompting often leads models to rely on superficial heuristics, resulting in a skewed initial probability distribution $p^{(0)}$.

\paragraph{Method.} We implement a two-step elicitation process. First, in \textit{Game Situation Reinterpretation}, the model is prompted to analyze the problem's structure and constraints explicitly. Second, in \textit{General Strategy Formulation}, it formulates high-level winning strategies. This step stops the model from locking into a wrong path early. We set the temperature to \texttt{0.1} to ensure consistent strategic output.

\paragraph{Theoretical Justification.} By forcing the model to ground its reasoning in domain-specific strategic knowledge, SPKE shifts the initial probability mass $p^{(0)}$ towards the optimal action. According to our model in Eq.~\ref{eq:prob_update}, increasing the initial probability of the correct argument is crucial because the subsequent debate dynamics tend to amplify the dominant initial belief.

\subsection{Stage 2: Perspective Diversification}
Even with a corrected initial belief, standard MAD suffers from an ``echo chamber'' effect where homogeneous agents reinforce each other's remaining biases. This corresponds to the persuasiveness being dependent on the argument's initial popularity rather than its quality.

\paragraph{Method.} We break this dependency by assigning each agent a unique, self-generated prompt. Inspired by the \textit{Learning from Multiple Approaches} theory~\citep{national2005students, cleaves2008promoting}, this encourages agents to adopt distinct reasoning paths. For example, as shown in Figure~\ref{fig:dreamad_process}, even starting from the same state, agents guided by diverse prompts generate distinct internal reasoning, preventing premature convergence to a wrong consensus. We set the temperature to \texttt{0.7} to promote diversity.

\paragraph{Theoretical Justification.} This module directly targets the debate dynamics. By forcing agents to adopt different perspectives, we ensure arguments win based on logic, not popularity $p^{(0)}$. This ensures that arguments are evaluated based on their intrinsic logical merit rather than their alignment with the majority bias, fostering a robust, truth-seeking debate.

After these stages, the agents conduct a structured multi-agent debate. The complete workflow is illustrated in Figure~\ref{fig:methods} and the detailed prompt formulation used in these two modules is documented in the Appendix \ref{app:sec:prompts}.
\section{Experiments}

\begin{figure*}[t]
\begin{center}
\centerline{\includegraphics[width=1.0\textwidth]{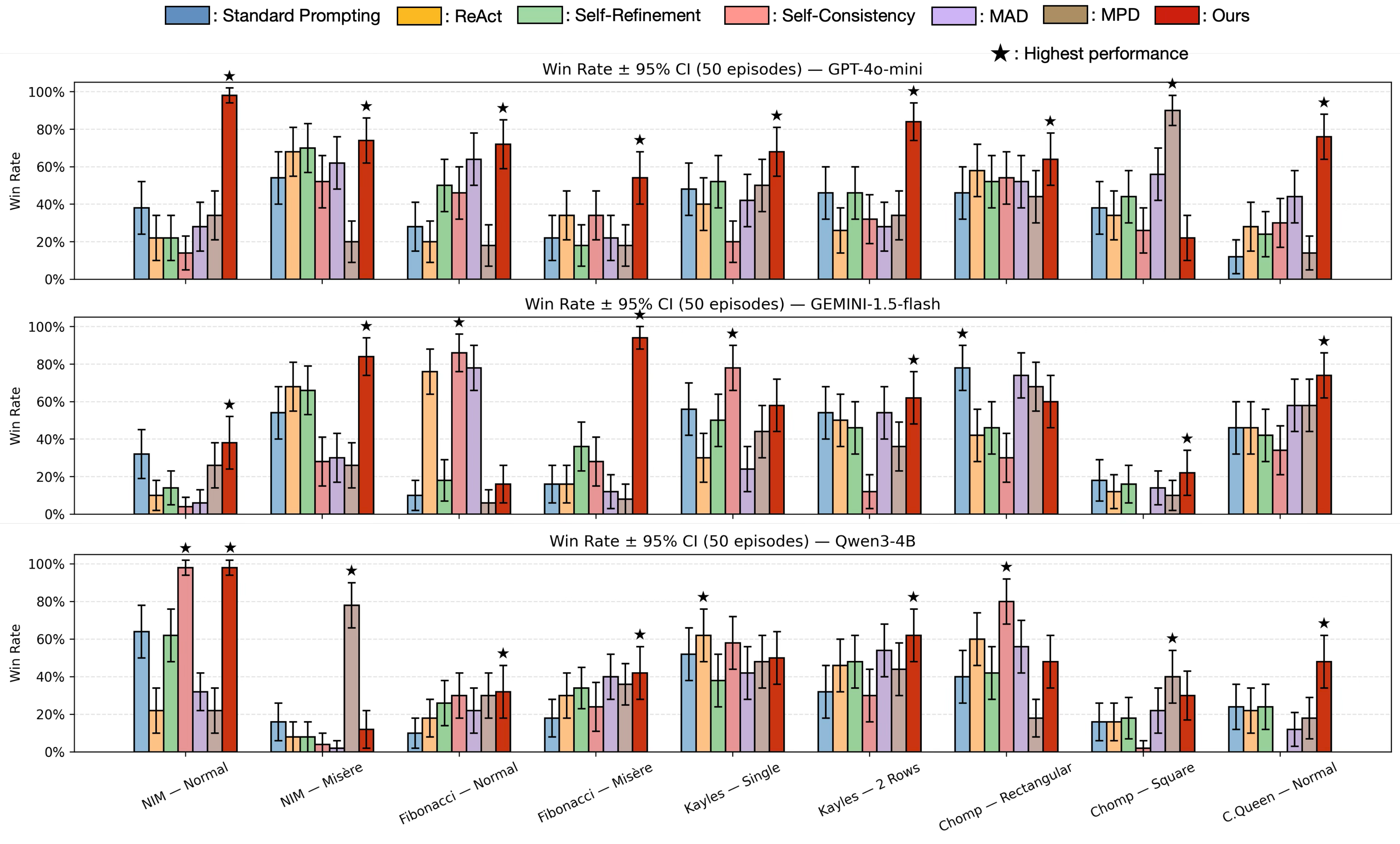}}
\vspace{-10pt}
\caption{Winning rate comparison across different models and different self-correction methods. This is a result based on \textit{MetaNIM Arena} simulator. The best results are highlighted in \textbf{bold}. We report win rates with 95\% confidence intervals over 50 episodes.}
\label{fig:main_results}
\end{center}
\vspace{-15pt}
\end{figure*}

We utilized the benchmark \textit{MetaNIM Arena} as both our dataset and simulator, as it provides a controlled environment for evaluating reasoning under grounded strategic tasks. Our investigation focuses on three key questions: (1) Does our approach improve reasoning quality compared to existing prompting techniques? (2) Does our approach prove its strategic reasoning quality under adversarial decision-making environments? (3) Does generating diverse prompts contribute to better decision-making within the debate framework? Further, we conduct two more ablation studies: (4) Can \alg achieve long CoT reasoning without further training? (5) Can \alg be generalized to general NLP domain?

To address these questions, we compare \alg with standard prompting methods including ReAct \citep{yao2023react}, Zero-shot Chain-of-Thought (CoT), Self-Consistency \citep{wang2022self}, Self-Refinement \citep{madaan2024self}, and Multi-Agent Debate (MAD \citep{du2023improving} and Multi-Persona Debate (MPD) \citep{liang2023encouraging}). The details of standard and CoT prompts are provided in Appendix \ref{app:sec:prompts}. 

Our method builds on the MAD framework by \citet{du2023improving}, augmenting it with structured self-prompt refinement and perspective diversification. For self-refinement, we follow the methodology of \citet{madaan2024self}, applying three iterative refinement steps. Similarly, for MAD, we conducted up to three rounds of debate, following \citet{du2023improving}, with the process terminating early if a consensus is reached before the final round. We also investigate whether the observed improvements generalize across different LLM architectures, including \texttt{GPT} \cite{achiam2023gpt}, \texttt{Gemini} \cite{team2023gemini}, and \texttt{Qwen} \cite{yang2025qwen3} models.

\subsection{Does \alg Improve Reasoning Quality?}
\vspace{-3pt}

This experiment isolates the effect of strategic prior knowledge elicitation, allowing us to assess whether our method enhances decision-making without relying on debate dynamics.

\noindent
\textbf{Setup.} To evaluate the effectiveness of our approach in improving reasoning capabilities, we compare \alg without the debate process against ReAct and Zero-shot CoT prompting across multiple models in the \textit{MetaNIM Arena} dataset (\S \ref{app:sec:dataset}). For showing versatility of \alg, we conduct experiments on five variants of LLMs as shown in Table~\ref{tab:reasoning_ability}.

\noindent
\textbf{Results.} 
Table~\ref{tab:reasoning_ability} demonstrates that \alg$^{\kern-0.3em(-)}$ generally improves performance over ReAct and CoT prompting across all models and tasks. These results highlight the impact of our method in reinforcing structured strategic reasoning, even without the iterative correction process of debate. Notably, our approach leads to substantial improvements in NIM, Fibonacci, and Kayles, which are environments where long-term strategic planning plays a crucial role. Since defining a general winning strategy in Chomp is non-trivial, applying prior knowledge is challenging and results in less effectiveness compared to other games. Furthermore, the gains tend to be larger for some weaker baselines (e.g., \texttt{GPT-4o-mini} by +17\% and \texttt{Gemini-1.5-flash} by +12\% on average), whereas very small models such as \texttt{Qwen3-4B} show little to no benefit, likely due to limited capacity of prompts refinement and eliciting priors.

\subsection{\alg in Adversarial Strategic Decision-Making}
\textbf{Setup.} We applied \alg to \texttt{GPT-4o-mini}, \texttt{Gemini-1.5-flash}, and \texttt{Qwen3-4B} then compared it with self-correction methods, including standard-prompt, ReAct, self-refinement, self-consistency, MAD and MPD. To demonstrate its effectiveness, we used \texttt{GPT-4o} as the opponent model due to its superior performance. In these experiments, we utilized \textit{MetaNIM Arena} simulator to maximize the effect of generating diverse prompts. We aimed to validate our hypothesis in a simulator that requires strategic decision-making within complex dynamics. We ran 50 independent episodes and average the win-rate.

\begin{figure}[t]
    \centering
    \includegraphics[width=1.0\columnwidth]{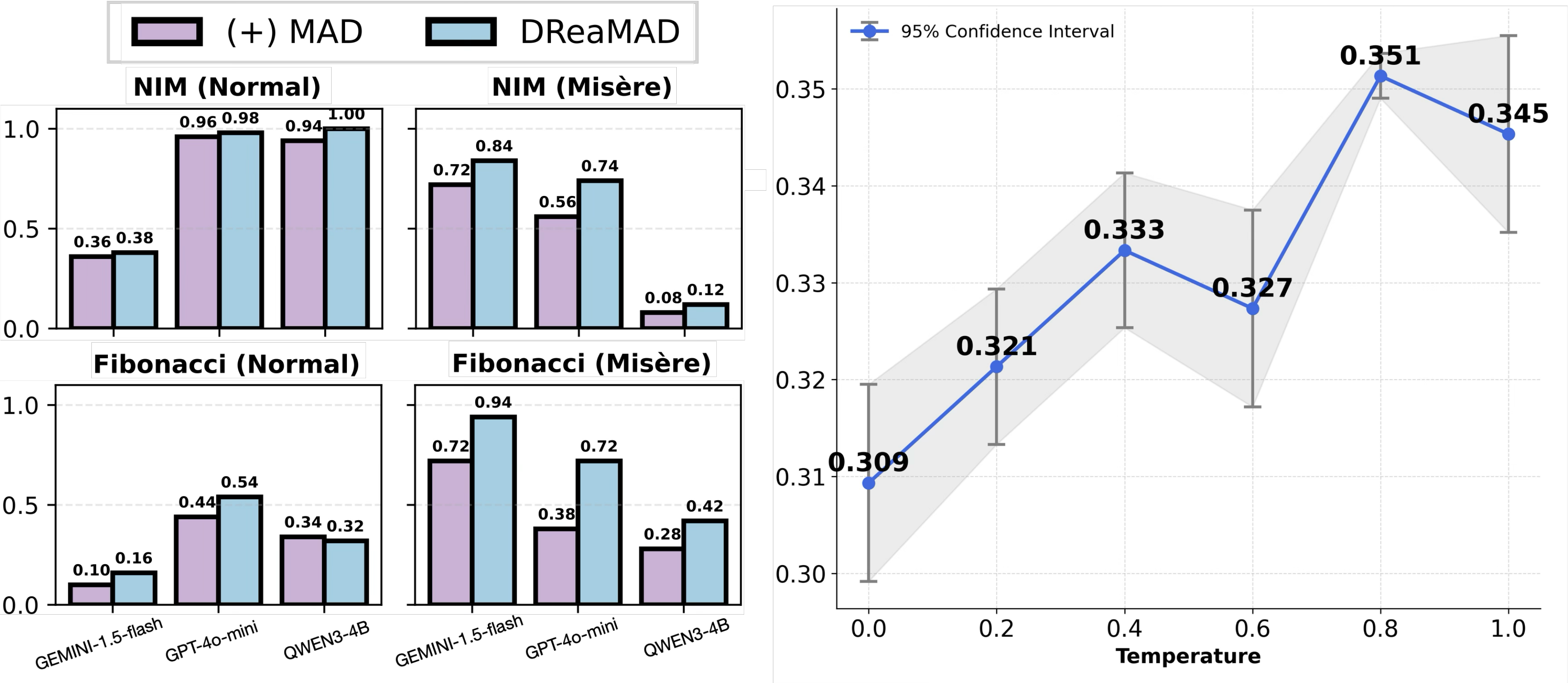} %
    \caption{Effect of perspective diversification. Left: Average win rate of \textbf{(+)MAD} and \alg on NIM and Fibonacci (Normal and Misère variants), aggregated over 50 simulations per setting. Right: Accuracy on the Fibonacci benchmark with \texttt{GPT-4o} across different sampling temperatures (15 runs each, 95\% CI). Higher temperatures yield greater prompt diversity, leading to improved accuracy.}
    \label{fig:diverse_prompts}
    \vspace{-10pt}
\end{figure}

\noindent
\textbf{Results.} As shown in Figure~\ref{fig:main_results}, \alg consistently outperforms other self-correction methods across various strategic environments, demonstrating a significant improvement in winning rates. This result suggests that our approach enables LLM agents to effectively adapt to complex dynamics, particularly in adversarial decision-making scenarios where strategic reasoning is crucial. Notably, even for a small model such as \texttt{Qwen3-4B}, \alg achieves the best overall performance among the compared methods; however, the magnitude of improvement tends to be larger for higher-capacity models, indicating that additional model capacity better translates elicited strategic priors into effective decision policies. However, we observe that \alg struggles in the Chomp game, which aligns with the fact that, although a winning strategy is guaranteed to exist for the first player by the strategy-stealing argument, no general closed-form solution is known for arbitrary board sizes. This proof establishes the existence of a winning strategy without explicitly constructing it, making the task highly non-trivial in practice. The absence of a fully characterized optimal strategy means that effective play often requires exploratory search rather than direct reasoning from prior knowledge. This highlights a limitation of our method in environments where strategic heuristics are only partially understood or insufficiently structured.

\subsection{Does Generating Diverse Prompts Improve Performance?}

We assess whether prompt diversity improves decision quality within the MAD framework. To this end, we compare two settings: (1) identical prompts generated via the Strategic Prior Knowledge Elicitation module for both agents as \textbf{(+)MAD}, and (2) distinct, self-generated prompts per agent, as in \alg (Figure~\ref{fig:diverse_prompts}, left). Experiments were conducted on four variants of the \textit{MetaNIM Arena} simulator, NIM (Normal and Misère) and Fibonacci (Normal and Misère), over 50 episodes each. Our results show that incorporating diverse prompts within the MAD framework significantly enhances performance, validating the effectiveness of our Perspective Diversification module.

We also examine the effect of sampling temperature on prompt diversity in the Fibonacci task. Within both the Strategic Prior Knowledge Elicitation and Perspective Diversification modules, we vary the temperature from 0.0 to 1.0. As shown in Figure~\ref{fig:diverse_prompts} right, higher temperature (further diversity) correlates with increased optimal action accuracy, indicating that greater diversity in generated prompts contributes to improved reasoning performance. However, we also note that excessively high temperatures could risk degrading reasoning coherence by introducing noise. Determining the optimal temperature that balances diversity and logical consistency remains an important consideration for practical applications.

\begin{table*}[ht]
\centering
\scriptsize
\setlength{\tabcolsep}{4pt}
\label{tab:math_tasks_unified}
\scalebox{0.92}{%
\begin{tabular}{@{}lcccccccccc@{}}
\toprule
\multirow{2}{*}{\textbf{Dataset}} &
\multicolumn{5}{c}{\texttt{GPT-o3-mini}} &
\multicolumn{5}{c}{\texttt{GPT-4o}} \\
\cmidrule(lr){2-6}\cmidrule(l){7-11}
& ReAct & Self-Refine. & Self-Consist. & MAD & \textbf{\alg}  %
& ReAct & Self-Refine. & Self-Consist. & MAD & \textbf{\alg} \\ %
\midrule
AIME 2024
& 72.0 \tiny $\pm$ 4.47 & 74.0 \tiny $\pm$ 1.49 & \textbf{82.7} \tiny $\pm$ 2.79 & \textbf{82.7} \tiny $\pm$ 2.79 & \textbf{82.7} \tiny $\pm$ 2.79
& 8.67 \tiny $\pm$ 4.47   & 8.67 \tiny $\pm$ 2.98  & 10.7 \tiny $\pm$ 2.71 & 11.3 \tiny $\pm$ 5.06  & \textbf{12.0} \tiny $\pm$ 5.58 \\
AMC 2023
& 99.0 \tiny $\pm$ 1.37  & 98.0 \tiny $\pm$ 1.12 & 99.5 \tiny $\pm$ 1.12 & 99.5 \tiny $\pm$ 1.12 & \textbf{100} \tiny $\pm$ 0.0
& 51.5 \tiny $\pm$ 3.29 & 54.5 \tiny $\pm$ 4.81 & 58.0 \tiny $\pm$ 5.42 & 55.0 \tiny $\pm$ 5.59 & \textbf{59.5} \tiny $\pm$ 4.47\\
\bottomrule
\end{tabular}%
}
\caption{Accuracy (\%) on math-reasoning benchmarks. We report the mean and standard deviation over 5 independent runs. Bold indicates the best algorithm for a given pair. While DReaMAD consistently achieves the highest mean accuracy, the overlap in standard deviations with Self-Consistency suggests comparable performance in some settings, highlighting the robustness of ensemble-based methods.}
\label{tab:math}
\end{table*}

\begin{figure}[t]
\vspace{-8pt}
    \centering
    \includegraphics[width=0.85\columnwidth]{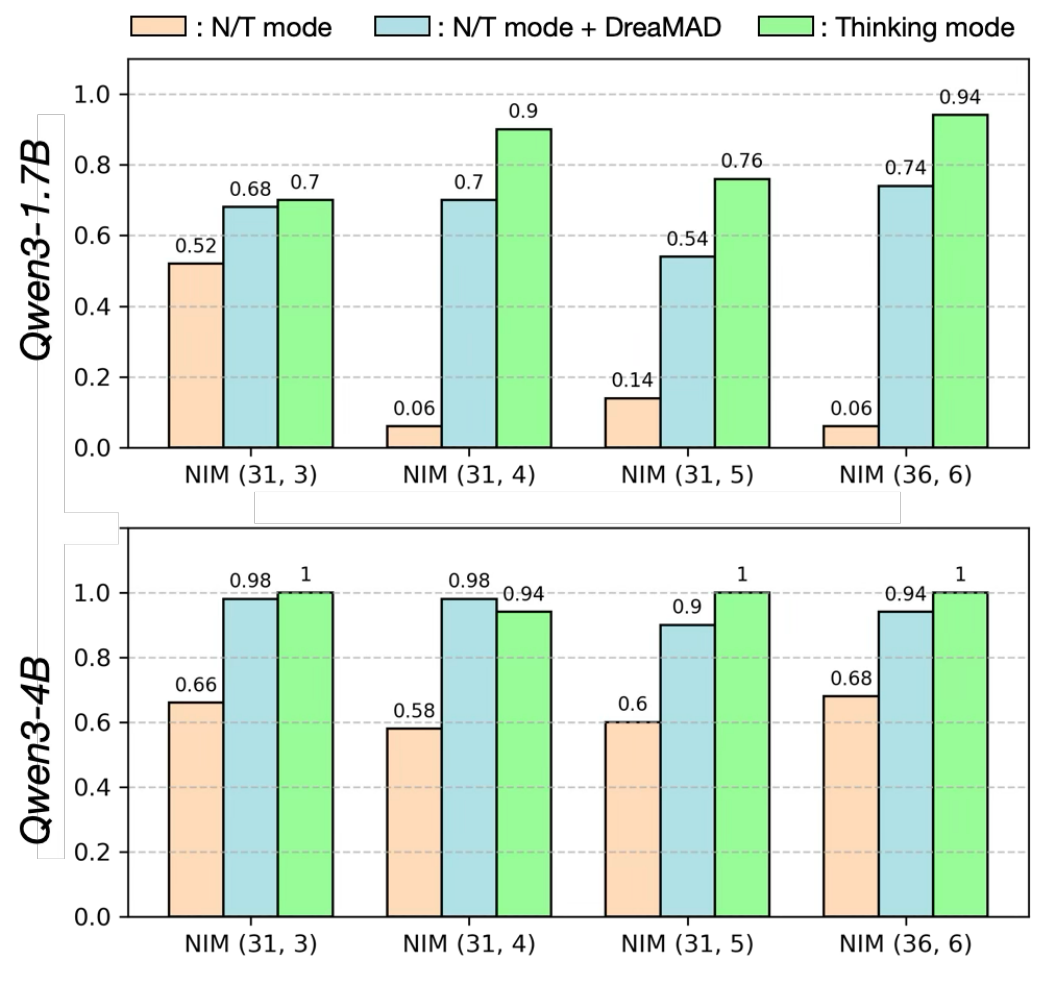} 
    \caption{Performance gain of \alg when applied to the Non-thinking (N/T) mode. The figure shows the average win rate over 50 runs for the \texttt{Qwen3-1.7B} (top) and \texttt{Qwen3-4B} (bottom) models. The notation NIM (n, k) indicates a game starting with n stones and a maximum take of k stones per turn.}
    \label{fig:qwen_thinking_mode}
    \vspace{-10pt}
\end{figure}

\subsection{Achieving Long CoT Reasoning Without Specialized Training}
\label{sec:qwen_analysis}

Researchers often assume that long CoT reasoning requires costly post-training, such as the multi-stage reinforcement learning pipeline used to create \texttt{Qwen3}'s thinking mode. In this section, we investigate a critical question: can the latent reasoning capabilities of a base model (non-thinking mode) be unlocked to match this long CoT reasoning performance without undergoing such training?

To explore this, we analyze the performance of the \texttt{Qwen3-1.7B} and \texttt{4B} model \citep{yang2025qwen3} on 4 variants of NIM-Normal game which have different initial number of stones and maximum number of stones that can be taken under three conditions, as shown in Figure~\ref{fig:qwen_thinking_mode}: (1) the standard, Non-thinking (N/T) mode; (2) the heavily post-trained Thinking mode; and (3) \alg on top of the N/T mode.

The results reveal the substantial value gained by our approach, demonstrating its effectiveness across different model scales. The standard Non-thinking mode models establish a performance baseline, which our framework significantly improves upon. For the \texttt{Qwen3-1.7B} model, \alg achieves a remarkable 226\% average performance gain, elevating a weak baseline to 81.3\% of the specialized Thinking mode's capability.

This effect is even more pronounced on the more capable \texttt{Qwen3-4B} model. Here, applying \alg yields a 45.1\% average performance gain, bringing the base model to 97.9\% of the Thinking mode’s performance and thus achieving near-parity: while the Thinking mode still attains the highest raw accuracy, \alg recovers most of its benefit without any additional post-training. The primary gain is one of efficiency and accessibility; \alg provides a training-free pathway to elicit long CoT reasoning from general-purpose models at test-time, bypassing the need for separate, resource-intensive training pipelines.

\subsection{Generalization to Math and Common Sense Reasoning}

\begin{table}[t]
\centering
\resizebox{0.98\linewidth}{!}{  
\begin{tabular}{@{}lccccc@{}}
\toprule
\textbf{Model} & ReAct & Self-Refine. &
Self-Consist. & MAD & \alg \\
\midrule
\texttt{GPT-4o}      & 83.93 \scriptsize $\pm$ 1.24 & 83.11 \scriptsize $\pm$ 1.37 & \textbf{84.75} \scriptsize $\pm$ 0.93 & 84.26 \scriptsize $\pm$ 1.47 & 83.93 \scriptsize $\pm$ 1.10 \\
\texttt{GPT-4o-mini} & 77.38 \scriptsize $\pm$ 1.37 & 75.25 \scriptsize $\pm$ 3.95& 76.89 \scriptsize $\pm$ 0.69 & 77.38 \scriptsize $\pm$ 0.73 & \textbf{78.85} \scriptsize $\pm$ 1.07\\
\bottomrule
\end{tabular}
}
\caption{Accuracy (\%) on CommonsenseQA dataset. We report the mean and standard deviation over 5 runs. Bold indicates the best performance. DReaMAD shows a statistically significant improvement over ReAct for \texttt{GPT-4o-mini}, while performing comparably to Self-Consistency for \texttt{GPT-4o}.}
\label{tab:commonsenseqa}
\end{table}

While our primary benchmark focuses on structured games, we further evaluate \alg on NLP tasks to test its broader applicability. Specifically, we consider algebra and number theory problems from AIME 2024 and AMC 2023, as well as CommonsenseQA~\citep{talmor2018commonsenseqa}. Experiments are conducted using standard accuracy metrics across model-task pairs, averaged over 5 independent runs.

Results in Table~\ref{tab:math} and Table~\ref{tab:commonsenseqa} show that \alg remains competitive or superior beyond our MetaNIM Arena benchmark.
In math reasoning (Table~\ref{tab:math}), \alg clearly improves over ReAct and Self-Refine. For example, on AIME 2024 with \texttt{GPT-o3-mini} it reaches 82.7\%, about 9–11 points higher than ReAct (72.0\%) and Self-Refine (74.0\%), and on AMC 2023 it attains 100\% accuracy, slightly surpassing Self-Consistency. For \texttt{GPT-4o}, \alg also gives the highest means on both AIME 2024 and AMC 2023, though the confidence intervals overlap with Self-Consistency and MAD, indicating comparable robustness. 

Similarly, in CommonsenseQA (Table~\ref{tab:commonsenseqa}), \alg achieves the highest mean accuracy for \texttt{GPT-4o-mini} (78.85\%), demonstrating statistically significant gains over baselines. For the stronger \texttt{GPT-4o} model, it performs on par with Self-Consistency. These findings align with our observations in Section~\ref{sec:qwen_analysis} (\texttt{Qwen} analysis): while specialized training (e.g., Thinking mode) sets a high baseline, \alg offers a training-free pathway to unlock latent reasoning capabilities, achieving long CoT performance or further refining the outputs of reasoning-specialized models.
\section{Related Works}
\label{sec:preliminary}

\subsection{Cognitive Bias and Belief Entrenchment in LLMs}
\label{sec:cognitive_bias}

While the term \textit{bias} in Large Language Models (LLMs) frequently refers to societal or demographic disparities~\citep{gallegos-etal-2024-bias}, our work investigates \textit{cognitive biases}, which are systematic deviations from rational decision-making that hinder robust reasoning. Specifically, we focus on \textbf{belief entrenchment}, a phenomenon where models exhibit excessive confidence in their initial, often erroneous, outputs and resist correction.

Recent studies suggest that LLMs often prioritize consistency over accuracy. This tendency arises because models are trained to maximize the likelihood of the next token based on context, leading them to align responses with user prompts or their own previous outputs rather than objective truth~\citep{perez-etal-2023-discovering}. Furthermore, \citet{turpin_2023} demonstrated that Chain-of-Thought (CoT) reasoning is frequently biased by the model's initial prediction. Instead of reasoning to find the answer, the model generates explanations to justify its initial choice. This aligns with our finding that models tend to reinforce their \textbf{static belief}, defined as the initial probability distribution over actions, regardless of its correctness.

This belief entrenchment is further intensified by the model's reliance on consistency as a proxy for correctness. While \citet{wang2022self} showed that aggregating consistent reasoning paths generally improves performance, this approach fails when the model is confident in a wrong answer, as shown. In such cases, the model consistently generates plausible but incorrect responses. Although models possess some capability to evaluate the validity of their own claims~\citep{kadavath2022languagemodelsmostlyknow}, this internal calibration often fails to override the reinforced consistency in iterative settings. Consequently, methods like Self-Refinement often fail to fix errors~\citep{huang2024large, valmeekam2022large}, instead creating ``reasoning loops'' that amplify the model's inherent biases. Our work extends these observations to the multi-agent setting, demonstrating that without structural intervention to diversify perspectives, Multi-Agent Debate (MAD) acts as a dynamic amplifier of these static cognitive biases, effectively forming an echo chamber of homogeneous agents.

\vspace{-5pt}
\subsection{Prompt Engineering and Self-Correction in LLMs}
\vspace{-2pt}

Prompt engineering shapes model outputs without retraining, potentially improving generalization and reducing bias~\citep{brown2020language, reynolds2021prompt, zhao-etal-2024-enhancing-zero, schulhoff2024promptreportsystematicsurvey, shin2024promptmodifierscontrolbias}. However, fully eliminating biases in complex reasoning remains challenging~\citep{jiang2020can, lu2022fantastically}. 

Meanwhile, self-correction mechanisms in LLMs refine responses without external supervision~\citep{ganguli2023capacitymoralselfcorrectionlarge, liu2024intrinsicselfcorrectioncapabilityllms, kamoi-etal-2024-llms}. Self-consistency, for instance, ensembles multiple outputs but converges on frequent rather than correct answers~\citep{wang2022self, chen2023universal}, and self-refinement can reinforce rather than fix biases~\citep{wan2023selfpolish, shinn2023reflexion, madaan2024self, huang2024large}. Feedback-loop methods such as STaR~\citep{zelikman2022star}, Reflexion~\citep{shinn2023reflexion}, and SCoRe~\citep{kumar2024traininglanguagemodelsselfcorrect} also struggle to reliably correct biases or foster diverse reasoning~\citep{guo2024biaslargelanguagemodels}.

\vspace{-5pt}
\subsection{Multi-Agent Debate in LLMs}
\label{sec:mad}
Multi-Agent Debate (MAD) enables LLM agents to critique each other, enhancing reasoning on complex tasks~\citep{liang2023encouraging,du2023improving}. ChatEval~\citep{chan2023chateval}, a multi-agent evaluation system, simulates human judgment to assess model output quality. Optimizations include task-specific strategies for improving debate effectiveness~\citep{smit2024shouldwemad} and ACC-Debate, an actor-critic framework that trains models to specialize in debates, achieving benchmark gains~\citep{estornell2024accdebateactorcritic}. Highlighting similar risks, \cite{estornell2024multi} provided a theoretical framework demonstrating that multi-LLM debates are susceptible to ``echo chamber" effects, where a flawed majority opinion arising from shared misconceptions can suppress correct minority viewpoints. While these enhancements improve performance, studies reveal a key limitation: static evaluations focus on assessing predefined problems, whereas real-world decision-making often involves dynamic, interactive environments where biases can evolve. Understanding how biases shift in these settings is crucial for developing robust strategies that extend beyond conventional static benchmarks. 
\section{Discussion}

\textbf{Limitation.}
While our approach demonstrates consistent improvements across a variety of strategic reasoning tasks, it is not without limitations. For instance, in the game of \textit{Chomp}, a winning strategy is guaranteed for the first player by the strategy-stealing argument, but no general, closed-form solution is known. Our framework, which relies on eliciting and refining prior strategic knowledge, is consequently less effective in such scenarios. This was reflected in our experiments, where performance on the Chomp dataset and simulator was significantly lower than on other games (Table~\ref{tab:reasoning_ability} and Figure~\ref{fig:main_results}). Chomp aside, we also observe that while our method improves performance on \textit{CommonsenseQA}, the gains are relatively modest compared to those on strategic and mathematical tasks (Table~\ref{tab:math}). This highlights that our method excels when a strong strategic foundation can be leveraged.

\noindent
\textbf{Future Directions.} A key future direction involves comparing our single-model approach with heterogeneous model ensembles. While employing different models could inherently increase perspective diversity, this often introduces significant practical overhead. Our work prioritizes a more tractable solution by eliciting diversity from a single model, a crucial step for real-world deployment. Systematically evaluating the trade-offs between these two strategies, and potentially integrating them, remains a promising area for future research.

\section{Conclusion}
Our study shows that Multi-Agent Debate (MAD) often reinforces biases instead of reducing them, leading to suboptimal reasoning. Through our experiments with the \textit{MetaNIM Arena}, we have observed that models persist in biased reasoning even when presented with superior alternatives.  While our current strategy focuses on strategic games,  the principles of structured self-refinement and diversified reasoning could be valuable for a wider range of NLP tasks. These include complex activities such as multi-step reasoning in question answering, legal analysis, and scientific inference.  Future work will explore how these techniques enhance decision-making beyond structured games.

\section*{Acknowledgments}
This work was supported by Center for Applied Research in Artificial Intelligence (CARAI) grant funded by Defense Acquisition Program Administration (DAPA) and Agency for Defense Development (ADD) (UD230017TD).

We sincerely thank the action editor, Antoine Bosselut, for their guidance and careful handling of the review process. We also thank the anonymous reviewers for their thoughtful and constructive feedback, which substantially improved the clarity and quality of this work.
\bibliography{tacl2021}

@article{du2023improving,
  title={Improving factuality and reasoning in language models through multiagent debate},
  author={Du, Yilun and Li, Shuang and Torralba, Antonio and Tenenbaum, Joshua B and Mordatch, Igor},
  journal={arXiv preprint arXiv:2305.14325},
  year={2023}
}

@article{liang2023encouraging,
  title={Encouraging divergent thinking in large language models through multi-agent debate},
  author={Liang, Tian and He, Zhiwei and Jiao, Wenxiang and Wang, Xing and Wang, Yan and Wang, Rui and Yang, Yujiu and Shi, Shuming and Tu, Zhaopeng},
  journal={arXiv preprint arXiv:2305.19118},
  year={2023}
}

@article{chan2023chateval,
  title={Chateval: Towards better llm-based evaluators through multi-agent debate},
  author={Chan, Chi-Min and Chen, Weize and Su, Yusheng and Yu, Jianxuan and Xue, Wei and Zhang, Shanghang and Fu, Jie and Liu, Zhiyuan},
  journal={arXiv preprint arXiv:2308.07201},
  year={2023}
}

@article{madaan2024self,
  title={Self-refine: Iterative refinement with self-feedback},
  author={Madaan, Aman and Tandon, Niket and Gupta, Prakhar and Hallinan, Skyler and Gao, Luyu and Wiegreffe, Sarah and Alon, Uri and Dziri, Nouha and Prabhumoye, Shrimai and Yang, Yiming and others},
  journal={Advances in Neural Information Processing Systems},
  volume={36},
  year={2024}
}

@article{wan2023selfpolish,
  title={Self-Polish: Enhance Reasoning in Large Language Models via Problem Refinement},
  author={Wan, Yuxuan and others},
  journal={arXiv preprint arXiv:2305.14497},
  year={2023}
}

@article{shinn2023reflexion,
  title={Reflexion: Language Agents with Verbal Reinforcement Learning},
  author={Shinn, Noah and Cassano, Federico and Berman, Edward and Gopinath, Ashwin and Narasimhan, Karthik and Yao, Shunyu},
  journal={arXiv preprint arXiv:2303.11366},
  year={2023}
}

@book{national2005students,
  title={How students learn},
  author={National Research Council and Donovan, Suzanne and Bransford, John and others},
  year={2005},
  publisher={National Academies Press Washington, DC}
}

@article{cleaves2008promoting,
  title={Promoting mathematics accessibility through multiple representations jigsaws},
  author={Cleaves, Wendy Pelletier},
  journal={Mathematics Teaching in the Middle School},
  volume={13},
  number={8},
  pages={446--452},
  year={2008},
  publisher={National Council of Teachers of Mathematics}
}

@misc{shin2024promptmodifierscontrolbias,
      title={Can Prompt Modifiers Control Bias? A Comparative Analysis of Text-to-Image Generative Models}, 
      author={Philip Wootaek Shin and Jihyun Janice Ahn and Wenpeng Yin and Jack Sampson and Vijaykrishnan Narayanan},
      year={2024},
      eprint={2406.05602},
      archivePrefix={arXiv},
      primaryClass={cs.CV},
}

@inproceedings{zhao-etal-2024-enhancing-zero,
    title = "Enhancing Zero-Shot Chain-of-Thought Reasoning in Large Language Models through Logic",
    author = "Zhao, Xufeng  and
      Li, Mengdi  and
      Lu, Wenhao  and
      Weber, Cornelius  and
      Lee, Jae Hee  and
      Chu, Kun  and
      Wermter, Stefan",
    editor = "Calzolari, Nicoletta  and
      Kan, Min-Yen  and
      Hoste, Veronique  and
      Lenci, Alessandro  and
      Sakti, Sakriani  and
      Xue, Nianwen",
    booktitle = "Proceedings of the 2024 Joint International Conference on Computational Linguistics, Language Resources and Evaluation (LREC-COLING 2024)",
    month = may,
    year = "2024",
    address = "Torino, Italia",
    publisher = "ELRA and ICCL",
    pages = "6144--6166",
}

@misc{schulhoff2024promptreportsystematicsurvey,
      title={The Prompt Report: A Systematic Survey of Prompting Techniques}, 
      author={Sander Schulhoff and Michael Ilie and Nishant Balepur and Konstantine Kahadze and Amanda Liu and Chenglei Si and Yinheng Li and Aayush Gupta and HyoJung Han and Sevien Schulhoff and Pranav Sandeep Dulepet and Saurav Vidyadhara and Dayeon Ki and Sweta Agrawal and Chau Pham and Gerson Kroiz and Feileen Li and Hudson Tao and Ashay Srivastava and Hevander Da Costa and Saloni Gupta and Megan L. Rogers and Inna Goncearenco and Giuseppe Sarli and Igor Galynker and Denis Peskoff and Marine Carpuat and Jules White and Shyamal Anadkat and Alexander Hoyle and Philip Resnik},
      year={2024},
      eprint={2406.06608},
      archivePrefix={arXiv},
      primaryClass={cs.CL},
}

@inproceedings{
huang2024large,
title={Large Language Models Cannot Self-Correct Reasoning Yet},
author={Jie Huang and Xinyun Chen and Swaroop Mishra and Huaixiu Steven Zheng and Adams Wei Yu and Xinying Song and Denny Zhou},
booktitle={The Twelfth International Conference on Learning Representations},
year={2024}
}

@article{wang2022self,
  title={Self-consistency improves chain of thought reasoning in language models},
  author={Wang, Xuezhi and Wei, Jason and Schuurmans, Dale and Le, Quoc and Chi, Ed and Narang, Sharan and Chowdhery, Aakanksha and Zhou, Denny},
  journal={arXiv preprint arXiv:2203.11171},
  year={2022}
}

@article{brown2020language,
  title={Language models are few-shot learners},
  author={Brown, Tom and Mann, Benjamin and Ryder, Nick and Subbiah, Melanie and Kaplan, Jared D and Dhariwal, Prafulla and Neelakantan, Arvind and Shyam, Pranav and Sastry, Girish and Askell, Amanda and others},
  journal={Advances in neural information processing systems},
  volume={33},
  pages={1877--1901},
  year={2020}
}

@article{achiam2023gpt,
  title={Gpt-4 technical report},
  author={Achiam, Josh and Adler, Steven and Agarwal, Sandhini and Ahmad, Lama and Akkaya, Ilge and Aleman, Florencia Leoni and Almeida, Diogo and Altenschmidt, Janko and Altman, Sam and Anadkat, Shyamal and others},
  journal={arXiv preprint arXiv:2303.08774},
  year={2023}
}

@article{dubey2024llama,
  title={The llama 3 herd of models},
  author={Dubey, Abhimanyu and Jauhri, Abhinav and Pandey, Abhinav and Kadian, Abhishek and Al-Dahle, Ahmad and Letman, Aiesha and Mathur, Akhil and Schelten, Alan and Yang, Amy and Fan, Angela and others},
  journal={arXiv preprint arXiv:2407.21783},
  year={2024}
}

@article{team2023gemini,
  title={Gemini: a family of highly capable multimodal models},
  author={Team, Gemini and Anil, Rohan and Borgeaud, Sebastian and Alayrac, Jean-Baptiste and Yu, Jiahui and Soricut, Radu and Schalkwyk, Johan and Dai, Andrew M and Hauth, Anja and Millican, Katie and others},
  journal={arXiv preprint arXiv:2312.11805},
  year={2023}
}

@article{chen2023universal,
  title={Universal self-consistency for large language model generation},
  author={Chen, Xinyun and Aksitov, Renat and Alon, Uri and Ren, Jie and Xiao, Kefan and Yin, Pengcheng and Prakash, Sushant and Sutton, Charles and Wang, Xuezhi and Zhou, Denny},
  journal={arXiv preprint arXiv:2311.17311},
  year={2023}
}

@misc{ganguli2023capacitymoralselfcorrectionlarge,
      title={The Capacity for Moral Self-Correction in Large Language Models}, 
      author={Deep Ganguli and Amanda Askell and Nicholas Schiefer and Thomas I. Liao and Kamilė Lukošiūtė and Anna Chen and Anna Goldie and Azalia Mirhoseini and Catherine Olsson and Danny Hernandez and Dawn Drain and Dustin Li and Eli Tran-Johnson and Ethan Perez and Jackson Kernion and Jamie Kerr and Jared Mueller and Joshua Landau and Kamal Ndousse and Karina Nguyen and Liane Lovitt and Michael Sellitto and Nelson Elhage and Noemi Mercado and Nova DasSarma and Oliver Rausch and Robert Lasenby and Robin Larson and Sam Ringer and Sandipan Kundu and Saurav Kadavath and Scott Johnston and Shauna Kravec and Sheer El Showk and Tamera Lanham and Timothy Telleen-Lawton and Tom Henighan and Tristan Hume and Yuntao Bai and Zac Hatfield-Dodds and Ben Mann and Dario Amodei and Nicholas Joseph and Sam McCandlish and Tom Brown and Christopher Olah and Jack Clark and Samuel R. Bowman and Jared Kaplan},
      year={2023},
      eprint={2302.07459},
      archivePrefix={arXiv},
      primaryClass={cs.CL}
}

@misc{kumar2024traininglanguagemodelsselfcorrect,
      title={Training Language Models to Self-Correct via Reinforcement Learning}, 
      author={Aviral Kumar and Vincent Zhuang and Rishabh Agarwal and Yi Su and John D Co-Reyes and Avi Singh and Kate Baumli and Shariq Iqbal and Colton Bishop and Rebecca Roelofs and Lei M Zhang and Kay McKinney and Disha Shrivastava and Cosmin Paduraru and George Tucker and Doina Precup and Feryal Behbahani and Aleksandra Faust},
      year={2024},
      eprint={2409.12917},
      archivePrefix={arXiv},
      primaryClass={cs.LG}
}

@article{cobbe2021training,
  title={Training verifiers to solve math word problems},
  author={Cobbe, Karl and Kosaraju, Vineet and Bavarian, Mohammad and Chen, Mark and Jun, Heewoo and Kaiser, Lukasz and Plappert, Matthias and Tworek, Jerry and Hilton, Jacob and Nakano, Reiichiro and others},
  journal={arXiv preprint arXiv:2110.14168},
  year={2021}
}

@article{edwards1994portable,
  title={Portable game notation specification and implementation guide},
  author={Edwards, Steven J},
  journal={Retrieved April},
  volume={4},
  pages={2011},
  year={1994}
}

@inproceedings{he2020box,
  title={The box is in the pen: Evaluating commonsense reasoning in neural machine translation},
  author={He, Jie and Wang, Tao and Xiong, Deyi and Liu, Qun},
  booktitle={Findings of the Association for Computational Linguistics: EMNLP 2020},
  pages={3662--3672},
  year={2020}
}

@article{kamoi-etal-2024-llms,
    title = "When Can {LLM}s Actually Correct Their Own Mistakes? A Critical Survey of Self-Correction of {LLM}s",
    author = "Kamoi, Ryo  and
      Zhang, Yusen  and
      Zhang, Nan  and
      Han, Jiawei  and
      Zhang, Rui",
    journal = "Transactions of the Association for Computational Linguistics",
    volume = "12",
    year = "2024",
    address = "Cambridge, MA",
    publisher = "MIT Press",
    doi = "10.1162/tacl_a_00713",
    pages = "1417--1440"
}

@misc{liu2024intrinsicselfcorrectioncapabilityllms,
      title={On the Intrinsic Self-Correction Capability of LLMs: Uncertainty and Latent Concept}, 
      author={Guangliang Liu and Haitao Mao and Bochuan Cao and Zhiyu Xue and Xitong Zhang and Rongrong Wang and Jiliang Tang and Kristen Johnson},
      year={2024},
      eprint={2406.02378},
      archivePrefix={arXiv},
      primaryClass={cs.CL}
}

@inproceedings{
zelikman2022star,
title={{ST}aR: Bootstrapping Reasoning With Reasoning},
author={Eric Zelikman and Yuhuai Wu and Jesse Mu and Noah Goodman},
booktitle={Advances in Neural Information Processing Systems},
editor={Alice H. Oh and Alekh Agarwal and Danielle Belgrave and Kyunghyun Cho},
year={2022}
}

@misc{guo2024biaslargelanguagemodels,
      title={Bias in Large Language Models: Origin, Evaluation, and Mitigation}, 
      author={Yufei Guo and Muzhe Guo and Juntao Su and Zhou Yang and Mengqiu Zhu and Hongfei Li and Mengyang Qiu and Shuo Shuo Liu},
      year={2024},
      eprint={2411.10915},
      archivePrefix={arXiv},
      primaryClass={cs.CL}
}

@inproceedings{reynolds2021prompt,
author = {Reynolds, Laria and McDonell, Kyle},
title = {Prompt Programming for Large Language Models: Beyond the Few-Shot Paradigm},
year = {2021},
isbn = {9781450380959},
publisher = {Association for Computing Machinery},
address = {New York, NY, USA},
doi = {10.1145/3411763.3451760},
booktitle = {Extended Abstracts of the 2021 CHI Conference on Human Factors in Computing Systems},
articleno = {314},
numpages = {7},
location = {Yokohama, Japan},
series = {CHI EA '21}
}

@misc{jiang2020can,
      title={Can Machines Learn Morality? The Delphi Experiment}, 
      author={Liwei Jiang and Jena D. Hwang and Chandra Bhagavatula and Ronan Le Bras and Jenny Liang and Jesse Dodge and Keisuke Sakaguchi and Maxwell Forbes and Jon Borchardt and Saadia Gabriel and Yulia Tsvetkov and Oren Etzioni and Maarten Sap and Regina Rini and Yejin Choi},
      year={2022},
      eprint={2110.07574},
      archivePrefix={arXiv},
      primaryClass={cs.CL}
}

@inproceedings{lu2022fantastically,
    title = "Fantastically Ordered Prompts and Where to Find Them: Overcoming Few-Shot Prompt Order Sensitivity",
    author = "Lu, Yao  and
      Bartolo, Max  and
      Moore, Alastair  and
      Riedel, Sebastian  and
      Stenetorp, Pontus",
    editor = "Muresan, Smaranda  and
      Nakov, Preslav  and
      Villavicencio, Aline",
    booktitle = "Proceedings of the 60th Annual Meeting of the Association for Computational Linguistics (Volume 1: Long Papers)",
    month = may,
    year = "2022",
    address = "Dublin, Ireland",
    publisher = "Association for Computational Linguistics",
    doi = "10.18653/v1/2022.acl-long.556",
    pages = "8086--8098"
}

@inproceedings{smit2024shouldwemad,
author = {Smit, Andries and Grinsztajn, Nathan and Duckworth, Paul and Barrett, Thomas D. and Pretorius, Arnu},
title = {Should we be going MAD? a look at multi-agent debate strategies for LLMs},
year = {2024},
publisher = {JMLR.org},
booktitle = {Proceedings of the 41st International Conference on Machine Learning},
articleno = {1866},
numpages = {23},
location = {Vienna, Austria},
series = {ICML'24}
}

@misc{estornell2024accdebateactorcritic,
      title={ACC-Debate: An Actor-Critic Approach to Multi-Agent Debate}, 
      author={Andrew Estornell and Jean-Francois Ton and Yuanshun Yao and Yang Liu},
      year={2024},
      eprint={2411.00053},
      archivePrefix={arXiv},
      primaryClass={cs.CL}
}

@inproceedings{yao2023react,
  title={React: Synergizing reasoning and acting in language models},
  author={Yao, Shunyu and Zhao, Jeffrey and Yu, Dian and Du, Nan and Shafran, Izhak and Narasimhan, Karthik and Cao, Yuan},
  booktitle={International Conference on Learning Representations (ICLR)},
  year={2023}
}

@article{sprague1935,
  author = {R. P. Sprague},
  title = {Über mathematische Kampfspiele},
  journal = {Tôhoku Mathematical Journal},
  volume = {41},
  pages = {438--444},
  year = {1935}
}

@article{grundy1939,
  author = {P. M. Grundy},
  title = {Mathematics and Games},
  journal = {Eureka},
  volume = {2},
  pages = {6--8},
  year = {1939}
}

@article{talmor2018commonsenseqa,
  title={Commonsenseqa: A question answering challenge targeting commonsense knowledge},
  author={Talmor, Alon and Herzig, Jonathan and Lourie, Nicholas and Berant, Jonathan},
  journal={arXiv preprint arXiv:1811.00937},
  year={2018}
}

@article{estornell2024multi,
  title={Multi-LLM debate: Framework, principals, and interventions},
  author={Estornell, Andrew and Liu, Yang},
  journal={Advances in Neural Information Processing Systems},
  volume={37},
  pages={28938--28964},
  year={2024}
}

@article{yang2025qwen3,
  title={Qwen3 technical report},
  author={Yang, An and Li, Anfeng and Yang, Baosong and Zhang, Beichen and Hui, Binyuan and Zheng, Bo and Yu, Bowen and Gao, Chang and Huang, Chengen and Lv, Chenxu and others},
  journal={arXiv preprint arXiv:2505.09388},
  year={2025}
}

@article{gallegos-etal-2024-bias,
    title = "Bias and Fairness in Large Language Models: A Survey",
    author = "Gallegos, Isabel O.  and
      Rossi, Ryan A.  and
      Barrow, Joe  and
      Tanjim, Md Mehrab  and
      Kim, Sungchul  and
      Dernoncourt, Franck  and
      Yu, Tong  and
      Zhang, Ruiyi  and
      Ahmed, Nesreen K.",
    journal = "Computational Linguistics",
    volume = "50",
    number = "3",
    month = sep,
    year = "2024",
    address = "Cambridge, MA",
    publisher = "MIT Press",
    url = "https://aclanthology.org/2024.cl-3.8/",
    doi = "10.1162/coli_a_00524",
    pages = "1097--1179"
}

@inproceedings{turpin_2023,
 author = {Turpin, Miles and Michael, Julian and Perez, Ethan and Bowman, Samuel},
 booktitle = {Advances in Neural Information Processing Systems},
 editor = {A. Oh and T. Naumann and A. Globerson and K. Saenko and M. Hardt and S. Levine},
 pages = {74952--74965},
 publisher = {Curran Associates, Inc.},
 title = {Language Models Don\textquotesingle t Always Say What They Think: Unfaithful Explanations in Chain-of-Thought Prompting},
 url = {https://proceedings.neurips.cc/paper_files/paper/2023/file/ed3fea9033a80fea1376299fa7863f4a-Paper-Conference.pdf},
 volume = {36},
 year = {2023}
}

@inproceedings{
valmeekam2022large,
title={Large Language Models Still Can't Plan (A Benchmark for {LLM}s on Planning and Reasoning about Change)},
author={Karthik Valmeekam and Alberto Olmo and Sarath Sreedharan and Subbarao Kambhampati},
booktitle={NeurIPS 2022 Foundation Models for Decision Making Workshop},
year={2022},
url={https://openreview.net/forum?id=wUU-7XTL5XO}
}

@inproceedings{perez-etal-2023-discovering,
    title = "Discovering Language Model Behaviors with Model-Written Evaluations",
    author = "Perez, Ethan  and
      Ringer, Sam  and
      Lukosiute, Kamile  and
      Nguyen, Karina  and
      Chen, Edwin  and
      Heiner, Scott  and
      Pettit, Craig  and
      Olsson, Catherine  and
      Kundu, Sandipan  and
      Kadavath, Saurav  and
      Jones, Andy  and
      Chen, Anna  and
      Mann, Benjamin  and
      Israel, Brian  and
      Seethor, Bryan  and
      McKinnon, Cameron  and
      Olah, Christopher  and
      Yan, Da  and
      Amodei, Daniela  and
      Amodei, Dario  and
      Drain, Dawn  and
      Li, Dustin  and
      Tran-Johnson, Eli  and
      Khundadze, Guro  and
      Kernion, Jackson  and
      Landis, James  and
      Kerr, Jamie  and
      Mueller, Jared  and
      Hyun, Jeeyoon  and
      Landau, Joshua  and
      Ndousse, Kamal  and
      Goldberg, Landon  and
      Lovitt, Liane  and
      Lucas, Martin  and
      Sellitto, Michael  and
      Zhang, Miranda  and
      Kingsland, Neerav  and
      Elhage, Nelson  and
      Joseph, Nicholas  and
      Mercado, Noemi  and
      DasSarma, Nova  and
      Rausch, Oliver  and
      Larson, Robin  and
      McCandlish, Sam  and
      Johnston, Scott  and
      Kravec, Shauna  and
      El Showk, Sheer  and
      Lanham, Tamera  and
      Telleen-Lawton, Timothy  and
      Brown, Tom  and
      Henighan, Tom  and
      Hume, Tristan  and
      Bai, Yuntao  and
      Hatfield-Dodds, Zac  and
      Clark, Jack  and
      Bowman, Samuel R.  and
      Askell, Amanda  and
      Grosse, Roger  and
      Hernandez, Danny  and
      Ganguli, Deep  and
      Hubinger, Evan  and
      Schiefer, Nicholas  and
      Kaplan, Jared",
    editor = "Rogers, Anna  and
      Boyd-Graber, Jordan  and
      Okazaki, Naoaki",
    booktitle = "Findings of the Association for Computational Linguistics: ACL 2023",
    month = jul,
    year = "2023",
    address = "Toronto, Canada",
    publisher = "Association for Computational Linguistics",
    url = "https://aclanthology.org/2023.findings-acl.847/",
    doi = "10.18653/v1/2023.findings-acl.847",
    pages = "13387--13434"
}

@misc{kadavath2022languagemodelsmostlyknow,
      title={Language Models (Mostly) Know What They Know}, 
      author={Saurav Kadavath and Tom Conerly and Amanda Askell and Tom Henighan and Dawn Drain and Ethan Perez and Nicholas Schiefer and Zac Hatfield-Dodds and Nova DasSarma and Eli Tran-Johnson and Scott Johnston and Sheer El-Showk and Andy Jones and Nelson Elhage and Tristan Hume and Anna Chen and Yuntao Bai and Sam Bowman and Stanislav Fort and Deep Ganguli and Danny Hernandez and Josh Jacobson and Jackson Kernion and Shauna Kravec and Liane Lovitt and Kamal Ndousse and Catherine Olsson and Sam Ringer and Dario Amodei and Tom Brown and Jack Clark and Nicholas Joseph and Ben Mann and Sam McCandlish and Chris Olah and Jared Kaplan},
      year={2022},
      eprint={2207.05221},
      archivePrefix={arXiv},
      primaryClass={cs.CL},
      url={https://arxiv.org/abs/2207.05221}, 
}
\bibliographystyle{acl_natbib}



\appendix
\clearpage

\section{MetaNIM Arena}
\label{app:metanim}

\begin{algorithm}
\caption{\textbf{MetaNIM Arena: } Turn-Based Opponent Task with Two Agents}
\label{app:alg:turn_based_task}
\begin{algorithmic}[1]
\REQUIRE Initial State $S_0$, Goal Condition $G$, Agents $A_1$ and $A_2$
\ENSURE Final State $S_f$ and Outcome
\STATE $t \gets 0$ \hfill \textit{Initialize turn counter}
\STATE $S \gets S_0$ \hfill \textit{Set initial state}
\WHILE{$S \not\in G$ \AND game is not terminated}
    \IF{$t \bmod 2 = 0$} 
        \STATE $a_t \gets A_1(S)$ \hfill \textit{Agent 1's turn, selects action $a_t$}
    \ELSE
        \STATE $a_t \gets A_2(S)$ \hfill \textit{Agent 2's turn, selects action $a_t$}
    \ENDIF
    \STATE $S \gets \text{UpdateState}(S, a_t)$ \hfill \textit{Apply the action and update state}
    \STATE $t \gets t + 1$ \hfill \textit{Increment turn counter}
    \IF{$S \in G$}
        \STATE \textbf{Success}: Goal Achieved
    \ENDIF
\ENDWHILE
\end{algorithmic}
\end{algorithm}

\subsection{Background on Impartial Games}
\label{app:sec:comb_games}

Games in the MetaNIM Arena are impartial games. We begin by outlining the relevant theoretical foundation, starting with the Sprague–Grundy theorem. In Table~\ref{app:tab:metanim_detailed}, we present the overall rules of the Nim game and its optimal strategies.

\subsubsection{Theory and Strategy}

\label{subsec:comb_games_theory}

All \textit{MetaNIM Arena} games are impartial games, forming a Directed Acyclic
Graph (DAG) where vertices represent game states and edges denote valid moves. The Grundy Number framework, along with the Sprague-Grundy Theorem, guarantees the existence of a winning strategy and provides a concrete method to determine it. In the \textit{MetaNIM Arena}, each state has a mathematically provable optimal move, allowing an LLM’s decisions to be evaluated against the \emph{theoretical optimal strategy}—a key advantage for unbiased assessment.

\begin{definition}[Grundy Number]
For a finite impartial game under normal play (where the last player to make a valid move wins), the \emph{Grundy number} (or \emph{Nimber}) $G(S)$ is recursively defined as follows. If $S$ is a \emph{terminal state} with no valid moves, set $G(S) = 0$. Otherwise,
\[
  G(S) \;=\; \mathrm{mex} \{\,G(S') \;|\; S'\!\text{ is reachable from }S\,\}\,.
\]
Here, $\mathrm{mex}(X)$ is the smallest nonnegative integer not in $X$. 
\end{definition}
Note that the Grundy number is well-defined for every impartial game, since the game’s state space forms a DAG. In many impartial games, direct enumeration of all possible move sequences is computationally infeasible. However, with Sprague-Grundy Theorem, we can easily calculate Grundy Numbers on complex games. See Appendix~\ref{app:sec:sprague-grundy} for further discussions.

\noindent\textbf{Optimal Strategy.}
When $G(S)\neq0$, there is \emph{at least one} move to a successor $S'$ with $G(S')=0$, forcing the
opponent into a losing position. Conversely, if $G(S)=0$, \emph{all} successor states have
$G(S')\neq0$. Because the game DAG is finite and acyclic, repeatedly applying “move to $G=0$” (or
avoiding it) ensures a \emph{forced} result under optimal play. See Appendix~\ref{app:sec:theory_strategy}
for more details, including the \emph{mis\`ere} variant where taking the last object \emph{loses}.

\subsubsection{Sprague-Grundy Theorem}
\label{app:sec:sprague-grundy}

The Sprague-Grundy theorem provides a fundamental method for analyzing impartial games by decomposing complex games into simpler, independent subgames. As discussed in Section~\ref{subsec:comb_games_theory}, every impartial game can be represented as a directed acyclic graph (DAG). However, directly computing Grundy numbers recursively from terminal states is often impractical.

We summarize key results from \citet{sprague1935} and \citet{grundy1939}. According to the theorem, the optimal strategy for playing multiple impartial games simultaneously (in parallel), or a single complex game viewed as multiple independent subgames, is equivalent to playing a single game of Nim with multiple heaps. This equivalence arises from the concept of the disjunctive sum of DAGs.

\begin{definition}[Disjunctive Sum of DAGs]
Let $\mathcal{G}_1 = (X_1,F_1), \mathcal{G}_2 = (X_2,F_2), \dots, \mathcal{G}_n = (X_n,F_n)$ be DAGs representing $n$ impartial games. The disjunctive sum of $\mathcal{G}_1, \dots, \mathcal{G}_n$ is a DAG $\mathcal{G} = (X,F)$ defined as follows:
\begin{enumerate}
    \item The vertex set $X$ is the Cartesian product $X_1 \times X_2 \times \dots \times X_n$.
    \item The edge set $F$ consists of edges connecting $(x_1,\dots,x_n)$ to $(y_1,\dots,y_n)$ if and only if exactly one pair $(x_i,y_i)$ is in $F_i$, and $x_j = y_j$ for all $j \neq i$.
\end{enumerate}

\textbf{Note.} In a disjunctive sum of DAGs, each player chooses exactly one subgame to play during their turn and moves within that subgame. The entire game ends when all subgames reach terminal positions.
\end{definition}

\begin{theorem}[Sprague-Grundy \citep{sprague1935,grundy1939}]
\label{thm:sprague_grundy}
A position $S$ is losing if and only if its Grundy number $G(S) = 0$; otherwise, if $G(S) \neq 0$, it is winning. Furthermore, if a position $S$ decomposes into independent subpositions $S_1, \dots, S_k$ via the disjunctive sum of DAGs, then
\[
  G(S)_{(2)} = G(S_1)_{(2)} \oplus G(S_2)_{(2)} \oplus \dots \oplus G(S_k)_{(2)},
\]
where $\oplus$ denotes bitwise XOR.
\end{theorem}

This result implies that Grundy numbers for complex games, such as Kayles or Chomp, can be efficiently computed by decomposing them into simpler subgames and combining the Grundy numbers using bitwise XOR. For example, consider a variant of the game Kayles played on two separate rows of pins, each forming an independent subgame. Suppose we computed the Grundy numbers separately for these rows, obtaining Grundy numbers $7$ for the first row and $4$ for the second row. By the Sprague-Grundy theorem, the combined Grundy number of the position is given by:
\[
  7_{(2)} \oplus 4_{(2)} = 111_{(2)} \oplus 100_{(2)} = 011_{(2)} = 3.
\]
Thus, even though the original game involves two distinct rows of pins, the strategic analysis reduces precisely to analyzing a Nim heap of size $3$. Since a Nim heap of size $3$ is nonzero, this indicates a winning position for the player about to move.

\subsubsection{Basic Discussions on the Optimal Strategy}
\label{app:sec:theory_strategy}
\paragraph{Why $G(S)=0$ Implies Losing.}
Recall that $G(S)$ is defined as:
\[
  G(S) \;=\; \mathrm{mex}\Bigl\{\,G(S') \,\Big|\; S'\!\text{ is reachable from } S\Bigr\},
\]
where $\mathrm{mex}(X)$ is the smallest nonnegative integer \emph{not} in the set $X$. Thus,
\[
  G(S)=0 \;\;\Longleftrightarrow\;\; 0 \notin \{G(S') \mid S' \in \mathcal{R}(S)\}
\]
where $\mathcal{R}(S)$ is the set of all states reachable from $S$. Concretely, if $G(S)=0$, then \emph{no valid move} leads to a successor $S'$ with $G(S')=0$. 
In other words, from $S$, the player to move \emph{cannot} transition the game into a $G(\cdot)=0$ state. 
Because a state $G(S')=0$ corresponds to a losing position \textit{for the player who faces it}, 
the mover in state $S$ has \emph{no way} to force the opponent into a losing position on the next turn.
Hence, $S$ is losing for the player to move.

\paragraph{Why $G(S)\neq 0$ Implies Winning (Opposite viewpoint).}
By the same logic, if $G(S)\neq 0$, then the definition of $\mathrm{mex}$ guarantees
$0$ \emph{does} appear among the Grundy values $G(S')$ of the successors. Thus, there \emph{exists}
some child state $S'$ for which $G(S')=0$. Consequently, the current mover can place the opponent
directly into a losing position (i.e.\ a position with Grundy number $0$). Recursively iterating this
argument along the Directed Acyclic Graph of states ensures that the current mover, if playing
optimally, keeps forcing the opponent into $G(\cdot)=0$ states until the game ends. Therefore,
$S$ must be a \emph{winning} state.

\noindent\textbf{Mis\`ere Variant.}
\emph{Mis\`ere} play reverses the normal condition: taking the last object loses rather than wins. Although standard Sprague-Grundy analysis still applies to most states, a special exception arises when all heaps (or subpositions) are size $1$, such as in \emph{mis\`ere} Nim. In that endgame scenario, the usual strategy must switch to avoid forcing the final move, ensuring the player leaves the opponent to pick the last object.

\begin{table*}[t]
\centering
\scriptsize 
\setlength{\tabcolsep}{6pt}
\renewcommand{\arraystretch}{1.15}
\begin{tabular}{@{}p{0.12\textwidth} p{0.38\textwidth} p{0.45\textwidth}@{}}
\toprule
\textbf{Game} & \textbf{Rule} & \textbf{Mathematical Strategy} \\
\midrule

\textbf{NIM} &
\begin{itemize}[leftmargin=*,nosep]
  \item $k$ heaps; remove from exactly one heap
  \item At least 1 up to a fixed maximum
  \item Last move wins (normal play)
\end{itemize}
&
\textbf{Core:} Nim-sum $= n_1 \oplus \cdots \oplus n_k$  
\begin{itemize}[leftmargin=*,nosep]
  \item \textbf{Winning:} Nim-sum $\neq 0$ (move to 0)  
  \item \textbf{Losing:} Nim-sum $=0$  
  \item Single-heap case $\Rightarrow$ modular arithmetic simplification
\end{itemize}
\\
\midrule

\textbf{Fibonacci Nim} &
\begin{itemize}[leftmargin=*,nosep]
  \item Start with $n$ items
  \item First move: $1 \leq k < n$
  \item Next: $1 \leq x \leq 2 \times$ (previous move)
  \item Last move wins
\end{itemize}
&
\textbf{Core:} Governed by Fibonacci numbers and Zeckendorf’s theorem  
\begin{itemize}[leftmargin=*,nosep]
  \item Zeckendorf: $m = F_{k_1}+F_{k_2}+\dots+F_{k_r}$, $|k_i-k_j|\geq 2$
  \item \textbf{Winning:} Reduce to nearest smaller Fibonacci $F_j$
  \item \textbf{Losing:} Pile size $=F_n$
  \item Optimal first move: remove the \emph{smallest} Fibonacci in decomposition
\end{itemize}
\\
\midrule

\textbf{Kayles} &
\begin{itemize}[leftmargin=*,nosep]
  \item Row of $n$ pins
  \item Knock down 1 pin or 2 adjacent pins
  \item Last move wins
\end{itemize}
&
\textbf{Core:} Grundy numbers via Sprague–Grundy theorem  
\begin{itemize}[leftmargin=*,nosep]
  \item Recurrence: $G(n) = \mathrm{mex}\{ G(n-1), G(n-2), G(a)\oplus G(b)\}$, $a+b=n-k$, $k\in\{1,2\}$
  \item \textbf{Winning:} Grundy $\neq 0$  
  \item \textbf{Losing:} Grundy $=0$; periodic mod 12 when $n\geq 70$
  \item Strategic principles: split into XOR-sum $=0$, mirror symmetric moves, avoid isolated single pins
\end{itemize}
\\
\midrule

\textbf{Chomp} &
\begin{itemize}[leftmargin=*,nosep]
  \item $m\times n$ chocolate grid, (1,1) poisoned
  \item Choose square; remove all above/right
  \item Last move wins
\end{itemize}
&
\textbf{Core:} Strategy-stealing $\Rightarrow$ first player always wins  
\begin{itemize}[leftmargin=*,nosep]
  \item \textbf{Winning:} Guaranteed for first player (though constructive proof unknown)  
  \item \textbf{Losing:} Not characterized except for small grids
  \item Analysis via posets under component-wise order
  \item Strategic principles: control antidiagonal, enforce symmetry, reduce to subgames, avoid isolated columns
\end{itemize}
\\
\midrule

\textbf{Corner Queen} &
\begin{itemize}[leftmargin=*,nosep]
  \item Queen at $(x,y)$ on $m\times n$ grid
  \item Move: left, down, or diagonal left-down by $k>0$
  \item Win by reaching $(0,0)$
\end{itemize}
&
\textbf{Core:} Equivalent to Wythoff’s Game; Beatty sequence structure  
\begin{itemize}[leftmargin=*,nosep]
  \item Losing ($P$-positions): $(\lfloor k\phi \rfloor, \lfloor k\phi^2 \rfloor)$, $\phi=\tfrac{1+\sqrt{5}}{2}$
  \item Examples: $(1,2), (3,5), (4,7), (6,10), (8,13), (9,15),\dots$
  \item \textbf{Winning:} Move to nearest $P$-position
\end{itemize}
\\
\bottomrule
\end{tabular}
\caption{MetaNIM Arena: Detailed Rules and Mathematical Strategies of Game Variants}
\label{app:tab:metanim_detailed}
\end{table*}

\subsection{Dataset and Simulator}
\label{app:sec:dataset}
We construct a simple dataset and a simulator based on the \textit{MetaNIM Arena}. The dataset focuses on specific scenes in each game and does not require an opponent model, while the simulator enables interaction with an opponent that receives prompts and outputs actions, where we employ the GPT-4o model as the opponent.

\begin{table}[ht]
\centering
\footnotesize
\vspace{5pt}
\resizebox{1.0\columnwidth}{!}{%
\begin{tabular}{@{}lcccc@{}}
\toprule[0.1em]
\textbf{Methods} & \textbf{NIM} & \textbf{Fibonacci} & \textbf{Chomp} & \textbf{Kayles} \\ 
\hline\hline
\multirow{2}{*}{Action space} & \multirow{2}{*}{1--3} & \multirow{2}{*}{dynamic (max 30)} & $x,y$ coordinate & single pin or \\
& & & (scenario-dependent) & two adjacent pins \\ \hline
Variants     & Normal & Normal & Square (2×2 – 19×19) & Normal \\ \hline
\# samples   & 20 & 11 & 20 & 18 \\ 
\bottomrule[0.1em]
\end{tabular}}
\caption{Constructed dataset using \textit{MetaNIM Arena}. We evaluate models on this dataset and report results in Table~\ref{tab:reasoning_ability}.} 
\label{app:tab:dataset}
\vspace{-0.2cm}
\end{table}

\begin{table}[ht]
\centering
\resizebox{1.0\columnwidth}{!}{ 
\begin{tabular}{lcccc}
\toprule
\multirow{2}{*}{\textbf{Features}}  & \multicolumn{2}{c}{\textbf{NIM}} & \multicolumn{2}{c}{\textbf{Fibonacci}} \\
\cmidrule(lr){2-3} \cmidrule(lr){4-5} 
& Normal & Misère & Normal & Misère \\
\midrule
\textbf{Starting Point} & remaining items: 31 & remaining items: 31 & remaining items: 20 & remaining items: 20 \\
\multirow{2}{*}{\textbf{Winning Condition}} & \multirow{2}{*}{taking last item} & \multirow{2}{*}{avoiding last item} & \multirow{2}{*}{taking last item} & \multirow{2}{*}{avoiding last item} \\
 & &  &  &  \\
\textbf{First Player?} & \cmark & \cmark & \cmark & \cmark \\
\textbf{Action Space} & 1 - 3 & 1 - 3 & dynamic & dynamic \\
\textbf{Opponent} & \multicolumn{4}{c}{\texttt{GPT-4o-2024-08-06}} \\
\bottomrule
\end{tabular}
}%
\caption{MetaNIM Arena Simulator: NIM and Fibonacci}
\label{app:tab:simulator1}
\end{table}

\begin{table}[ht]
\centering
\resizebox{1.0\columnwidth}{!}{%
\begin{tabular}{lccccc}
\toprule
\multirow{2}{*}{\textbf{Features}} &
\multicolumn{2}{c}{\textbf{Kayles}} &
\multicolumn{2}{c}{\textbf{Chomp}} &
\textbf{Corner Queen} \\  

\cmidrule(lr){2-3} \cmidrule(lr){4-5} \cmidrule(lr){6-6} 

& Single & 2 Rows & Rectangular & Square & Normal \\    
\midrule
\textbf{Starting Point}          & remaining items: 20 & piles 5\!+\!6 & 2×8 & 5×5 & queen at $(4,16)$ \\
\multirow{2}{*}{\textbf{Winning Condition}}
                                 & \multirow{2}{*}{take last item} & \multirow{2}{*}{take last item}
                                 & avoid poison & avoid poison & reach $(0,0)$ \\
                                 &                          &                          & (top-left)   & (top-right)  & (lower-left corner) \\
\textbf{First Player?}           & \cmark & \cmark & \cmark & \cmark & \cmark \\
\textbf{Action Space}            & pile index & pile index (row, column) & $x,y$ coordinate & $x,y$ coordinate & $x,y$ coordinate \\
\textbf{Opponent}                & \multicolumn{5}{c}{\texttt{GPT-4o-2024-08-06}} \\
\bottomrule
\end{tabular}
}%
\caption{MetaNIM Arena Simulator: Kayles, Chomp, and Corner Queen}
\label{app:tab:simulator2}
\end{table}

\subsubsection{Game Prompts}
We provide how our MetaNIM Arena game's (NIM, Fibonacci, Chomp, Kayles, Corner-Queen) prompts are constructed as shown in Table~\ref{app:tab:game_prompts}. Any users can modify the games as they want to make.

\begin{table*}[t!]
\centering
\footnotesize
\resizebox{0.85\textwidth}{!}{%
\begin{tabularx}{\textwidth}{@{} >{\bfseries}p{1.9cm} >{\RaggedRight\arraybackslash}X @{}}
\toprule
\textbf{Game} & \textbf{Game Prompt} \\
\midrule

Standard & 
\textbf{\#Game Role:} You are \{agent[`name']\}, a participant in a game of Nim variants.\par\vspace{3pt}
\textbf{\#Objective:} Your goal is to win the game by taking all remaining items on your turn, leaving no items for your opponent. The person who takes the last item wins.\par\vspace{3pt}
\textbf{\#Game Rule:} There is a single pile of items. You can take between 1 and \{max\_take\} items on your turn.\par\vspace{3pt}
\textbf{\#Current State:} There are \{remaining\_items\} items remaining in the pile.\par\vspace{3pt}
\textbf{\#Task:} Based on the current state of the game, decide how many items you will take (between 1 and \{max\_take\}) on this turn.\par\vspace{3pt}
\textbf{Output Format:} The output should be a Markdown code snippet with the following scheme, including leading and trailing triple backticks with \texttt{"json"} and:\par\vspace{3pt}
\texttt{```}\texttt{\{}\texttt{action: integer // This is an action you take. Only integer between 1 and 3.}\texttt{\}}
\texttt{'''}\par\vspace{3pt}\\

ReAct & 
\textbf{Output Format:}
The output should be a Markdown code snippet with the following scheme, including leading and trailing triple backticks with \texttt{"json"} and:\par\vspace{3pt}
\texttt{```}\texttt{\{}\texttt{reasoning: string // This is the reason for the action}\par\vspace{3pt}
\texttt{action: integer // This is an action you take based on the reasoning. Only integer between 1 and 3.}\texttt{\}}
\texttt{'''}\par\vspace{3pt} \\

CoT & 
\textbf{Output Format:}
The output should be a Markdown code snippet with the following scheme, including leading and trailing triple backticks with \texttt{"json"} and:\par\vspace{3pt}
\texttt{```}
\texttt{\{}
\texttt{reasoning: string // This is the reason for the action}\par\vspace{3pt}
\texttt{action: integer // This is an action you take based on the reasoning. Only integer between 1 and 3.}
\texttt{\}}
\texttt{'''}\par\vspace{3pt}
Let's think step-by-step. What is the best move for you?\par\vspace{3pt} \\

\alg & 
\textbf{\#Game Situation Reinterpretation:}\par\vspace{3pt}
\texttt{game\_prompt} : \par\vspace{3pt}
\texttt{Below is a game description. Extract the key information.}\par\vspace{3pt}
\texttt{Game Description: \{current state\}}\par\vspace{3pt}
\texttt{\#\#\# Format response as:}\par\vspace{3pt}
\texttt{```}\texttt{\{}
\texttt{game definition: string // What is the definition of this game?}\par\vspace{3pt}
\texttt{winning condition: string // How to win the game.}\par\vspace{3pt}
\texttt{move constraints: string // What actions are allowed per turn.}
\texttt{\}}
\texttt{'''}\par\vspace{3pt}

\textbf{\#General Strategy Formulation:}\par\vspace{3pt}
\texttt{strategy\_prompt} : \par\vspace{3pt}
\texttt{Based on the game information below, derive the general winning strategy in this game}\par\vspace{3pt}
\texttt{Game: \{game definition\}}\par\vspace{3pt}
\texttt{Winning Condition: \{winning condition\}}\par\vspace{3pt}
\texttt{Move Constraints: \{move constraints\}}\par\vspace{3pt}
\texttt{Current State: \{current state in very short\}}\par\vspace{3pt}
\texttt{\#\#\# Format response as:}\par\vspace{3pt}
\texttt{```}\texttt{\{}
\texttt{state evaluation: string // How to assess the game state.}\par\vspace{3pt}
\texttt{winning strategy: string // Winning strategy in this turn to win this game.}\par\vspace{3pt}
\texttt{endgame tactics: string // Best strategy in a near-win situation.}
\texttt{\}}\texttt{'''}\par\vspace{3pt}

\textbf{\#Perspective Diversification:}\par\vspace{3pt}
\texttt{Refine the initial game prompt to improve decision-making based on the Game and Strategy Information.}

\texttt{Initial Prompt: \{given initial prompt\}}\par\vspace{3pt}
\texttt{Game and Strategy Information:}\par\vspace{3pt}
\texttt{Game: \{game definition\}}\par\vspace{3pt}
\texttt{Strategy:}\par\vspace{3pt}
\texttt{- State Evaluation: \{state evaluation\}}\par\vspace{3pt}
\texttt{- Winning Strategy: \{winning strategy\}}\par\vspace{3pt}
\texttt{- Endgame Tactics: \{endgame tactics\}}\par\vspace{3pt}
\texttt{\#\#\# Format response as:}\par\vspace{3pt}
\texttt{```}\texttt{\{}
\texttt{optimized prompt: string // The refined prompt that clearly directs decision-making.}\texttt{\}}
\texttt{'''}\par\vspace{3pt}
\\

\bottomrule
\end{tabularx}
}
\caption{Standard, ReAct, CoT, \alg Prompt in NIM game.}
\label{app:tab:all_reasoning_prompts}
\end{table*}

\begin{table*}[t!]
\centering
\footnotesize
\begin{tabularx}{\textwidth}{@{} >{\bfseries}p{1.9cm} >{\RaggedRight\arraybackslash}X @{}}
\toprule
\textbf{Game} & \textbf{Game Prompt} \\
\midrule

NIM & 
\textbf{\#Game Role}: You are \{agent['name']\}, a participant in a game of Nim variants. \par\vspace{3pt}
\textbf{\#Objective}: Your goal is to win the game by taking all remaining items on your turn, leaving no items for your opponent. The person who takes the last item wins. \par\vspace{3pt}
\textbf{\#Game Rule}: There is a single pile of items. You can take between 1 and \{max\_take\} items on your turn. \par\vspace{3pt}
\textbf{\#Current State}: There are \{remaining\_items\} items remaining in the pile. \par\vspace{3pt}
\textbf{\#Task}: Based on the current state of the game, decide how many items you will take (between 1 and \{max\_take\}) on this turn. \\
\addlinespace

Fibonacci & 
\textbf{\#Game Role}: You are \{agent['name']\}, a participant in a simple Fibonacci game. \par\vspace{3pt}
\textbf{\#Objective}: Your goal is to win the game by taking all remaining stones on your turn, leaving no stones for your opponent. The person who takes the last stones wins. \par\vspace{3pt}
\textbf{\#Game Rule}: 1. There is a single pile of stones. \newline 2. Players take turns one after another. \newline 3. The first player can take any number of stones, but not all the stones in the first move. \newline 4. On subsequent turns, the number of stones a player can take must be at least 1 and at most twice the number of stones the previous player took. \newline 5. The player who takes the last stone wins the game. \par\vspace{3pt}
\textbf{\#Current State}: There are \{remaining\_items\} stones remaining in the pile. \par\vspace{3pt}
\textbf{\#Task}: You are the first player. Based on the current state of the game, decide how many items you will take (between 1 and \{remaining\_items - 1\}) on this turn. \\
\addlinespace

Chomp & 
\textbf{\#Game Role}: You are \{agent['name']\}, a participant in a game of Chomp. \par\vspace{3pt}
\textbf{\#Objective}: Your goal is to force your opponent to take the top-left corner of the grid (position (0, 0)). \par\vspace{3pt}
\textbf{\#Game Rule}: 1. The game is played on a square grid. \newline 2. On your turn, you select a position (row, col). \newline 3. All positions to the right and below the selected position are removed. \newline 4. The player forced to select (0, 0) loses. \par\vspace{3pt}
\textbf{\#Current State}: The grid is represented as a binary matrix, where '1' means the position is still available, and '0' means it is removed: \{remaining\_grid\} \par\vspace{3pt}
\textbf{\#Task}: Based on the current state of the grid, decide which position (row, col) you will select. \\
\addlinespace

Kayles &
\textbf{\#Game Role}: You are \{agent['name']\}, a participant in a game of Kayles. \par\vspace{3pt}
\textbf{\#Objective}: Your goal is to win the game by leaving your opponent with no valid moves. The player who takes the last pin(s) wins. \par\vspace{3pt}
\textbf{\#Game Rule}: 1. There is a single row of pins. \newline 2. On your turn, you can remove: \newline \textbullet~1 pin, \newline \textbullet~2 adjacent pins. \newline 3. You cannot remove non-adjacent pins or pins that have already been removed. \par\vspace{3pt}
\textbf{\#Current State}: The row of pins is represented as a binary string: \newline – '1' means the pin is still there. \newline – '0' means the pin has already been removed. \newline Current state: \{remaining\_pins\} \par\vspace{3pt}
\textbf{\#Task}: Based on the current state of the game, decide which pin(s) you will take on this turn. \\
\addlinespace

Corner-Queen &
\textbf{\#Game Role}: You are \{agent['name']\}, a participant in a Corner-Queen game. \par\vspace{3pt}
\textbf{\#Objective}: Move the queen so that \textbf{you} are the first to place it on the lower-left corner square. \par\vspace{3pt}
\textbf{\#Game Rule}: 1. Board size: \{board\_height\}$\times$\{board\_width\}. \newline 2. Coordinates use zero-based indices [row, col]. Row 0 is the top row; Col 0 is the leftmost column. Valid ranges: row $\in$ [0, board\_height - 1], col $\in$ [0, board\_width - 1]. \newline 3. From the current position [r, c] the queen may move to **one** of: (a) left: [r, c'] with c' < c; (b) down: [r', c] with r' > r; (c) left-down diagonal: [r + d, c - d] with d > 0. \newline 4. The game ends when the queen reaches [row = board\_height-1, col = 0]. \par\vspace{3pt}
\textbf{\#Current State}: Current position: [row = \{r\}, col = \{c\}]. \par\vspace{3pt}
\textbf{\#Task}: Based on the current state, decide the next move [row, col]. \\

\bottomrule
\end{tabularx}
\caption{Game state input prompts.}
\label{app:tab:game_prompts}
\end{table*}

\subsubsection{Reasoning and \alg Prompts}
\label{app:sec:prompts}
\paragraph{Standard, ReAct \& CoT Prompts.}
In our evaluation of LLMs within the \textit{MetaNIM Arena,} we compare several key prompting techniques: Standard, Zero-shot Chain-of-Thought (CoT), ReAct Prompting and \alg as shown in Table \ref{app:tab:all_reasoning_prompts}. The distinction between these approaches significantly impacts the model’s reasoning and decision-making process.

\paragraph{Prompting Strategy for Opponent Modeling.}
Anything can act as an opponent in the \textit{MetaNIM Arena} simulator, but we model OpenAI’s \texttt{GPT-4o}, the most powerful LLM model currently available, as the opponent and apply the ReAct prompting method.

\clearpage

\section{Ablation study}
\label{app:sec:ablation_setup}

\begin{table*}[t]
\centering
\begin{minipage}{0.48\textwidth}
    \centering
    \footnotesize
    \resizebox{1.0\textwidth}{!}{%
    \begin{tabular}{lcccccc}
    \toprule[0.1em]
    \multirow{2.3}{*}{\textbf{(a)} \texttt{GPT-4o}}
    & \multicolumn{3}{c}{\textbf{Wrong Belief}} & \multicolumn{3}{c}{\textbf{Good Belief}} \\
    \cmidrule(lr){2-4} \cmidrule(lr){5-7} 
    & (20, 19) & (12, 4) & (7, 4) & (15, 10) & (16, 8) & (7, 7) \\
    \midrule
    Standard & 0.700 & 0.675 & 0.600 & 0.525 &  0.725 & 0.850 \\
    \cellcolor{gray!20}+ After MAD & \cellcolor{gray!20}\textbf{0.900} & \cellcolor{gray!20}\textbf{0.750} & \cellcolor{gray!20}\textbf{0.700} & \cellcolor{gray!20}\textbf{0.750} & \cellcolor{gray!20}\textbf{0.750} & \cellcolor{gray!20}\textbf{0.900} \\[-0.18em]
    \bottomrule[0.1em]
    \end{tabular}
    }
    \label{tab:gpt4o_bias}
\end{minipage}
\hfill
\begin{minipage}{0.48\textwidth}
    \centering
    \footnotesize
    \resizebox{1.0\textwidth}{!}{%
    \begin{tabular}{lcccccc}
    \toprule[0.1em]
    \multirow{2.3}{*}{\textbf{(b)} \texttt{GPT-4o-mini}}  
    & \multicolumn{4}{c}{\textbf{Wrong Belief}} & \multicolumn{2}{c}{\textbf{Good Belief}} \\
    \cmidrule(lr){2-5} \cmidrule(lr){6-7} 
    & (18, 4) & (12, 6) & (10, 4) & (15, 10) & (7, 2) & (15, 2) \\
    \midrule
    Standard & 0.700 & 0.875 & 0.600 & 0.975 & 0.950 & 0.975 \\
    \cellcolor{gray!20}+ After MAD & \cellcolor{gray!20}\textbf{0.850} & \cellcolor{gray!20}\textbf{0.950} & \cellcolor{gray!20}\textbf{0.750} & \cellcolor{gray!20}\textbf{1.000} & \cellcolor{gray!20}0.950 & \cellcolor{gray!20}\textbf{1.000} \\[-0.18em]
    \bottomrule[0.1em]
    \end{tabular}
    }
    \label{tab:gpt4omini_bias}
\end{minipage}

\vspace{0.5em}

\begin{minipage}{0.48\textwidth}
    \centering
    \footnotesize
    \resizebox{1.0\textwidth}{!}{%
    \begin{tabular}{lcccccc}
    \toprule[0.1em]
    \multirow{2.3}{*}{\textbf{(c)} \texttt{Gemini-1.5-pro}}  
    & \multicolumn{3}{c}{\textbf{Wrong Belief}} & \multicolumn{3}{c}{\textbf{Good Belief}} \\
    \cmidrule(lr){2-4} \cmidrule(lr){5-7} 
    & (20, 19) & (12, 4) & (4, 4) & (15, 10) & (10, 4) & (16, 8) \\
    \midrule
    Standard & 0.650 & 0.600 & 0.600 & 0.800 & 0.625 & 0.675 \\
    \cellcolor{gray!20}+ After MAD & \cellcolor{gray!20}\textbf{0.700} & \cellcolor{gray!20}\textbf{0.650} & \cellcolor{gray!20}\textbf{0.800} & \cellcolor{gray!20}0.800 & \cellcolor{gray!20}\textbf{0.800} & \cellcolor{gray!20}\textbf{0.700} \\[-0.18em]
    \bottomrule[0.1em]
    \end{tabular}
    }
    \label{tab:geminipro_bias}
\end{minipage}
\hfill
\begin{minipage}{0.48\textwidth}
    \centering
    \footnotesize
    \resizebox{1.0\textwidth}{!}{%
    \begin{tabular}{lcccccc}
    \toprule[0.1em]
    \multirow{2.3}{*}{\textbf{(d)} \texttt{Gemini-1.5-flash}}  
    & \multicolumn{3}{c}{\textbf{Wrong Belief}} & \multicolumn{3}{c}{\textbf{Good Belief}} \\
    \cmidrule(lr){2-4} \cmidrule(lr){5-7} 
    & (12, 6) & (12, 4) & (7, 4) & (15, 4) & (20, 19) & (7, 7) \\
    \midrule
    Standard & 0.800 & 0.700 & 0.525 & 0.750 & 0.500 & 0.750 \\
    \cellcolor{gray!20}+ After MAD & \cellcolor{gray!20}\textbf{0.850} & \cellcolor{gray!20}\textbf{1.000} & \cellcolor{gray!20}\textbf{0.750} & \cellcolor{gray!20}0.750 & \cellcolor{gray!20}\textbf{0.650} & \cellcolor{gray!20}\textbf{1.000} \\[-0.18em]
    \bottomrule[0.1em]
    \end{tabular}
    }
    \label{tab:geminiflash_bias}
\end{minipage}

\caption{Belief entrenchment across models: showing that even after the debate concludes, strongly consistent actions continue to exhibit strong consistency, reinforcing initial static belief in the Fibonacci game. A wrong belief occurs when the model’s biased response deviates from the optimal action, while a correct belief refers to cases where the wrong dominant response aligns with the optimal action. (\textit{a, b, c, d}) indicate the state where the remaining items are \textit{a}, and player can take the items maximum to \textit{b}.}
\vspace{-5pt}
\label{tab:fibonacci_analysis}
\end{table*}

\subsection{Instance of Belief Entrenchment}
\label{app:sec:instance_of_bias}

\begin{figure}[t]
\vspace{-15pt}
    \includegraphics[width=0.5\textwidth]{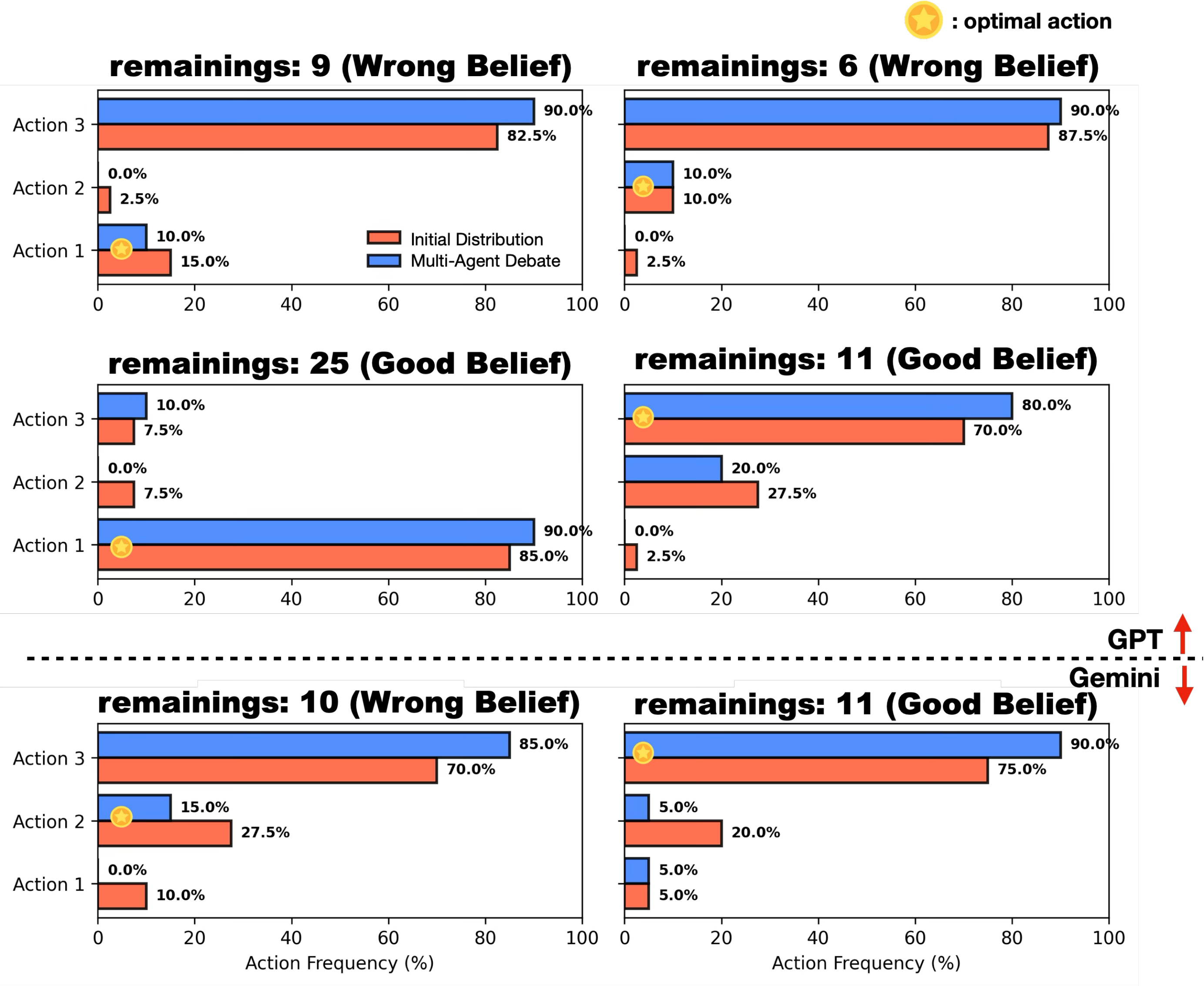}
    \caption{Belief entrenchment in NIM game by MAD. We compared initial action distribution and action distribution after 3 rounds of debates.}
    \label{fig:nim_analysis}
    \vspace{-15pt}
\end{figure}

\noindent
\textbf{In NIM:} In Figure~\ref{fig:nim_analysis}, we find that MAD amplifies models’ pre-existing biases rather than refining their reasoning. To investigate this, we first identify game states where each model exhibits \textit{strong consistency}, i.e., consistently selecting the same action across multiple trials. For each such state, we conduct MAD using two identical agents instantiated from the same model. Each agent generates 20 responses (40 per state) at a fixed temperature of 0.7, maintained throughout the debate. Initial action distributions (light red) are compared against post-debate distributions after three rounds (blue). If MAD functioned as a self-correction mechanism, we would expect the distribution to shift toward the optimal action.

However, we found the opposite: regardless of whether the initial reasoning was correct, the debate process consistently amplifies pre-existing beliefs rather than mitigating them. For instance, in Figure~\ref{fig:nim_analysis} (top-left), \texttt{GPT-4o-mini} initially selects a suboptimal action (Action 3) 82.5\% of the time. After the debate, this frequency increases to 90.0\%, while the proportion of optimal responses drops further. Rather than correcting errors, the debate reinforces strongly consistent—but incorrect—responses. This pattern persists across model families. The Gemini models (bottom row of Figure~\ref{fig:nim_analysis}) exhibit similar behavior, regardless of whether the initial belief aligns with optimal play (“good belief”) or not (“wrong belief”). In both cases, MAD strengthens the dominant trajectory without introducing new strategic insight.

Figure~\ref{fig:debate_process} provides a detailed view: two LLM agents receive the same input and begin debating. Despite initial divergence, they converge quickly—after the first round—on a shared line of reasoning. Crucially, this convergence occurs even when the initial consensus is incorrect, illustrating that MAD often serves as an amplifier of belief rather than a correction mechanism.

\textbf{In Fibonacci:} Extending our NIM analysis, we evaluate MAD’s effect in the Fibonacci game, a more complex setting with move constraints and dynamic interactions. As in NIM, we identify states exhibiting strong consistency and categorize them into two groups: those where consistent responses align with the optimal strategy, and those where they do not. We then apply MAD to examine how response distributions evolve post-debate. Consistent with the NIM results, MAD reinforces the model’s initial beliefs in Fibonacci as well. As shown in Table~\ref{tab:fibonacci_analysis}, the frequency of initially consistent actions increases after debate, regardless of whether those actions are optimal. On average, reinforcement rises by 9.17\% in GPT models and 12.29\% in Gemini models, indicating a systematic amplification of dominant reasoning patterns
across architectures.

\begin{table}[t]
\centering
\small
\setlength{\tabcolsep}{6pt}
\scalebox{0.90}{%
\begin{tabular}{@{}lcccccc@{}}
\toprule
\textbf{Metric} & \textbf{1 ep.} & \textbf{2 ep.} & \textbf{3 ep.} & \textbf{4 ep.} & \textbf{\alg{}} \\
\midrule
Win-rate            & 0.253 & 0.300 & 0.420 & 0.460 & \textbf{0.966} \\
API cost (\$)       & 0.013 & 0.023 & 0.033 & 0.043 & \textbf{0.0098} \\
\bottomrule
\end{tabular}
}
\caption{Performance and API expenditure for
\texttt{GPT-4o-mini} fine-tuned on \textsc{NIM-N} (1–4 epochs)
versus our zero-training \alg inference. Values are averaged over three game variants (50 matches each).}
\label{tab:nim_cost}
\vspace{-4pt}
\end{table}

\subsection{Cost effectiveness of \alg}
While \alg requires a single model, its inference involves additional prompt steps (e.g., prior knowledge elicitation) and a debate process. However, this test-time scaling method is much more efficient than train-time scaling. As we utilized language models through an API, we compared the costs using dollar amounts, contrasting the API cost per single game under our method with the costs of OpenAI fine-tuning models on constructed datasets for NIM-N games.

The NIM-N dataset is built from three different variants of the NIM game. Each game starts with 31 stones, but the maximum removable stones per turn differs (3, 4, or 5). For each variant, eight distinct game states were sampled with their optimal actions as labels, yielding 24 training examples in total.

Additionally, testing was conducted on the aforementioned three scenarios (each 50 games) by using the \texttt{GPT-4o} model as the opponent. The win rate was measured for each scenario, and the average win rate across these variants was reported.

The results of this comparison are illustrated in the table above. When fine-tuning a model using API-based fine-tuning (\texttt{GPT-4o-mini}), the performance gradually improved with additional training epochs, achieving win-rates of 0.253, 0.300, 0.420, and 0.460 at 1, 2, 3, and 4 epochs respectively, with corresponding API costs of \$0.013, \$0.023, \$0.033, and \$0.043 (Here, the cost of constructing dataset is not included). In contrast, our proposed method, \alg, achieved significantly higher performance (0.966) with substantially lower API costs (\$0.0098). These results strongly suggest that our approach not only outperforms traditional fine-tuning methods but also is far more cost-efficient. All prices are calculated by the pricing policy: \href{https://openai.com/api/pricing/}{https://openai.com/api/pricing/}

\begin{table}[t]
\vspace{-10pt}
\centering
\small
\setlength{\tabcolsep}{5pt}
\scalebox{0.75}{%
\begin{tabular}{@{}lcccc@{}}
\multicolumn{5}{c}{\texttt{GPT-4o-mini}} \\[2pt]
\toprule
\textbf{Method} & \textbf{NIM-N} & \textbf{NIM-M} & \textbf{Fib-N} & \textbf{Fib-M} \\
\midrule
Self-Refinement                       & 0.22 & 0.70 & 0.18 & 0.50 \\
Self-Refinement + \alg$^{\kern-0.3em(-)}$        & 0.66 & 0.66 & 0.16 & 0.46 \\
Self-Consistency                      & 0.14 & 0.52 & 0.34 & 0.46 \\
Self-Consistency + \alg$^{\kern-0.3em(-)}$       & 0.34 & 0.66 & 0.48 & 0.54 \\
MAD                & 0.28 & 0.62 & 0.22 & 0.82 \\
\textbf{\alg}                    & \textbf{0.98} & \textbf{0.74} & \textbf{0.54} & \textbf{0.94} \\
\bottomrule
\end{tabular}
}
\vspace{0.8em}  

\scalebox{0.75}{%
\begin{tabular}{@{}lcccc@{}}
\multicolumn{5}{c}{\texttt{Gemini-1.5-flash}} \\[2pt]
\toprule
\textbf{Method} & \textbf{NIM-N} & \textbf{NIM-M} & \textbf{Fib-N} & \textbf{Fib-M} \\
\midrule
Self-Refinement                       & 0.14 & 0.66 & 0.18 & 0.36 \\
Self-Refinement + \alg$^{\kern-0.3em(-)}$        & 0.34 & 0.80 & 0.10 & 0.30 \\
Self-Consistency                      & 0.04 & 0.28 & \textbf{0.28} & 0.86 \\
Self-Consistency + \alg$^{\kern-0.3em(-)}$       & \textbf{0.80} & 0.54 & 0.18 & 0.30 \\
MAD               & 0.06 & 0.30 & 0.12 & 0.78 \\
\textbf{\alg}                    & 0.38 & \textbf{0.84} & 0.16 & \textbf{0.94} \\
\bottomrule
\end{tabular}
}
\caption{Win-rates on four impartial game variants when applied \alg$^{\kern-0.3em(-)}$ to Self-Refinement and Self-Consistency method. \textbf{Bold} numbers mark the best score in each column.  
\alg$^{\kern-0.3em(-)}$ indicates ablations where debate is not applied during inference.}
\label{tab:game_results}
\vspace{-10pt}
\end{table}

\begin{figure*}[ht]
\vskip 0.2in
\begin{center}
\centerline{\includegraphics[width=1.0\textwidth]{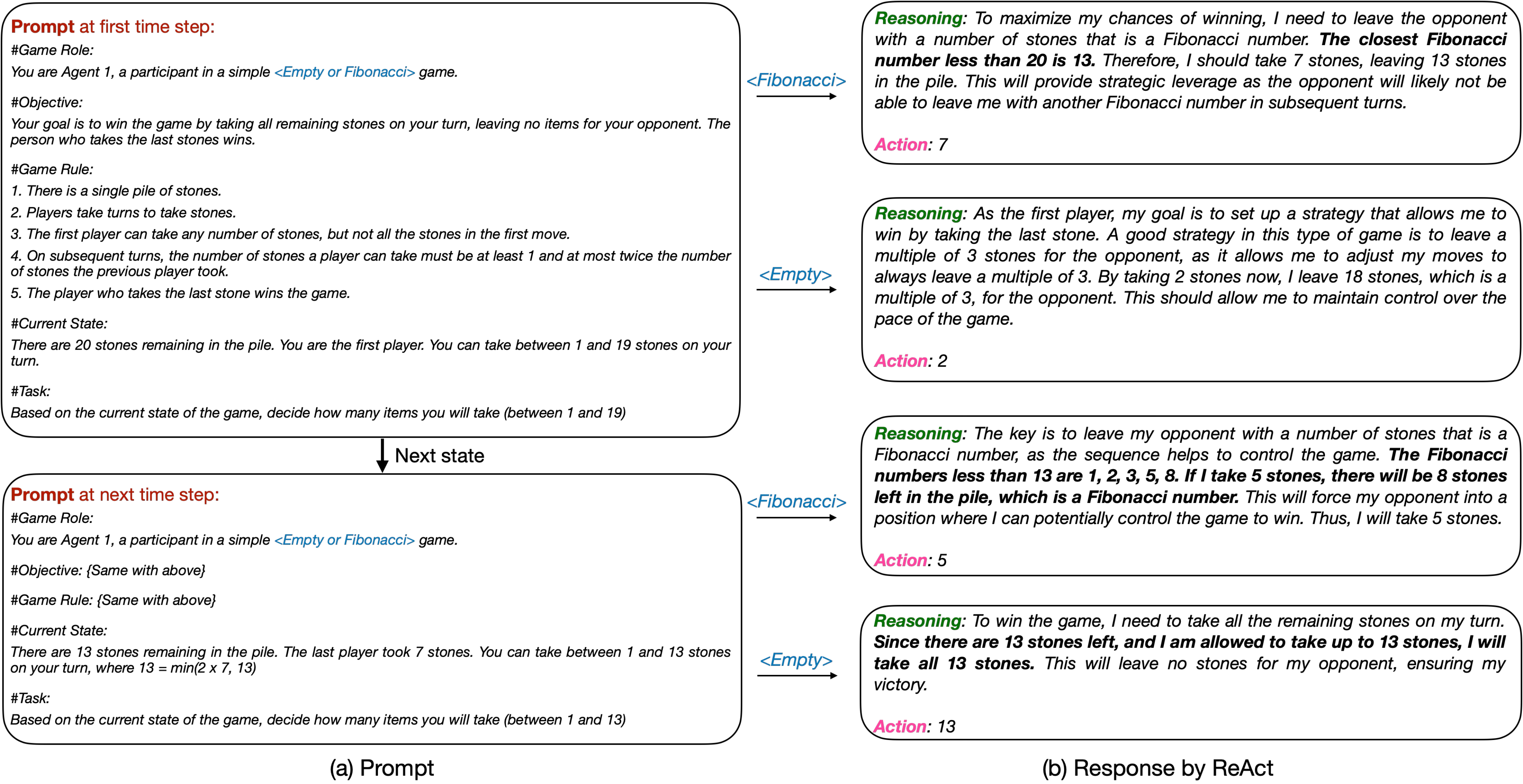}}
\caption{Reasoning comparison based on the presence of the keyword `Fibonacci' in the prompt. Explicit usage aligns reasoning with optimal strategy, demonstrating the impact of lexical cuing on strategic planning.}
\label{app:fig:prompt_bias}
\end{center}
\vskip -0.2in
\end{figure*}

\subsection{Applicability of Diverse Amplification on Self-Reflection and Self-Consistency}
We show whether other self-correction methods, such as Self-Refinement and Self-Consistency, can benefit from structured guidance that enhances reasoning diversity. In \alg, this is operationalized through the Strategic Prior Knowledge Elicitation (SPKE) module, which prompts the model to reinterpret the problem and formulate general strategies before engaging in debate. To isolate SPKE’s impact, we evaluate \alg, which includes SPKE but excludes debate (see Table~\ref{tab:reasoning_ability}). We further apply SPKE to Self-Refinement and Self-Consistency and compare them to their vanilla versions. The results show that SPKE alone consistently improves performance across settings.

\subsection{Belief Entrenchment in NIM Scenario}
\label{app:nim_case_study}

For Figure~\ref{fig:motif_example}, we describe the specific scenario where debate facilitates belief entrenchment. 
The setup is a NIM-Normal game with 5 items remaining. An agent must decide how many items to take. With \emph{Strong Consistency}, an agent evaluates possible moves on its own and concludes that taking 2 items (leaving 3 for the opponent) would lead to a winning position. Conducting MAD, two agents exchange their reasoning before deciding. An agent suggests taking 2 items but provides flawed reasoning, while another argues that taking 1 item is better since it leaves the opponent with 4 items, a multiple of 4, which is known to be strategically advantageous.

\subsection{A Word Change in Prompts Leads to Different Output}
\label{app:sec:word_change}

Figure~\ref{app:fig:prompt_bias} compares LLM responses when the word \textit{Fibonacci} is explicitly mentioned versus when it is omitted in an identical game scenario. In the presence of the keyword \textit{Fibonacci}, the model aligns its reasoning with Fibonacci-based strategy, leveraging number sequences to maintain control over the game. Conversely, when the term is absent, the model defaults to an alternative heuristic, such as maintaining a multiple of three or even resorting to a trivial greedy strategy. For instance, in the first decision step, when instructed with \textit{Fibonacci}, the model identifies 13 as the closest Fibonacci number and takes 7 stones, ensuring an advantageous future state. Without the keyword, however, the model applies a modulo-based heuristic, taking only 2 stones to leave a multiple of three. Similarly, in the second decision step, the \textit{Fibonacci}-aware model deliberately leaves 8 stones in the pile—another Fibonacci number—while the other instance simply takes all remaining stones without strategic foresight.

\end{document}